%% file: main.tex
\documentclass[10pt,journal,compsoc]{IEEEtran}
\ifCLASSOPTIONcompsoc
  \usepackage[nocompress]{cite}
\else
  \usepackage{cite}
\fi

\usepackage{epsfig}
\usepackage{epstopdf}
\usepackage{graphicx}
\usepackage{amsmath}
\usepackage{bm}
\usepackage{amssymb}
\usepackage{color, colortbl}
\usepackage{xcolor}
\usepackage[english]{babel}

\usepackage{arydshln}

\usepackage{algorithm} 
\usepackage{algpseudocode} 
\usepackage{booktabs}
\usepackage{verbatim}
\usepackage{gensymb}
\usepackage{tikz}
\usepackage{pgffor}

\usepackage{xspace}
\usepackage{multirow}
\usepackage{enumitem}
\usepackage{rotating}
\usepackage{makecell}
\usepackage{marvosym}

\usepackage{ragged2e}

\usepackage[normalem]{ulem}
\usepackage[breaklinks=true,colorlinks,bookmarks=false]{hyperref}

\usepackage{pifont}

\RequirePackage{fontawesome}

\makeatletter
\DeclareRobustCommand\onedot{\futurelet\@let@token\@onedot}
\def\@onedot{\ifx\@let@token.\else.\null\fi\xspace}

\def\eg{{e.g}\onedot} 
\def\ie{{i.e}\onedot} 
 
\def\etc{{etc}\onedot} 
\def\wrt{w.r.t\onedot} 
\def\etal{{et al}\onedot}
\makeatother

\makeatletter
\let\MYcaption\@makecaption
\makeatother

\usepackage[font=footnotesize,labelformat=simple]{subcaption}

\makeatletter
\let\@makecaption\MYcaption
\makeatother

\makeatletter
\def\zapcolorreset{\let\reset@color\relax\ignorespaces}
\def\colorrows#1{\noalign{\aftergroup\zapcolorreset#1}\ignorespaces}
\makeatother

\renewcommand{\paragraph}[1]{\textbf{#1}}

\begin{document}

\title{PyMAF-X: Towards Well-aligned Full-body Model Regression from Monocular Images}

\author{Hongwen Zhang,
        Yating Tian,
        Yuxiang Zhang,
        Mengcheng Li,
        Liang An,
        Zhenan Sun,~\IEEEmembership{Senior Member,~IEEE},
        and Yebin Liu,~\IEEEmembership{Member,~IEEE}
\IEEEcompsocitemizethanks{\IEEEcompsocthanksitem Hongwen Zhang, Yuxiang Zhang, Mengcheng Li, Liang An, and Yebin Liu are with the Department of Automation, Tsinghua University, Beijing 100084, China.
E-mail: \{zhanghongwen,liuyebin\}@mail.tsinghua.edu.cn; \{yx-z19,li-mc18,al17\}@mails.tsinghua.edu.cn;
(Corresponding author: Yebin Liu)
\IEEEcompsocthanksitem Yating Tian is with the Department of Computer Science and Technology, Nanjing University, Nanjing 210023, China.
E-mail: \mbox{yatingtian@smail.nju.edu.cn}
\IEEEcompsocthanksitem Zhenan Sun is with the Institute of Automation, Chinese Academy of Sciences, Beijing 100190, China.
E-mail: \mbox{znsun@nlpr.ia.ac.cn}.
}}

\markboth{IEEE Transactions on Pattern Analysis and Machine Intelligence}%
{Zhang \MakeLowercase{\textit{et al.}}: PyMAF-X: Towards Well-aligned Full-body Model Regression from Monocular Images}

\input{tex/0_Abstract}

\maketitle

\IEEEdisplaynontitleabstractindextext
\IEEEpeerreviewmaketitle

\input{tex/1_Introduction}

\input{tex/2_RelatedWork}

\input{tex/3_Methodology}
\input{tex/4_Experiments}

\input{tex/5_Conclusion}

\vspace{-3mm}

\ifCLASSOPTIONcompsoc
  \section*{Acknowledgments}
\else
  \section*{Acknowledgment}
\fi

This work was supported by the National Key R\&D Program of China (2021ZD0113501), the National Natural Science Foundation of China (No.62125107, U1836217, and 62276263), and the China Postdoctoral Science Foundation (No.2022M721844).
We would like to thank Xinchi Zhou, Wanli Ouyang, and Limin Wang for their help, feedback, and discussions in the early work of this paper.

\ifCLASSOPTIONcaptionsoff
  \newpage
\fi

\bibliographystyle{IEEEtran}
\bibliography{IEEEabrv,bibtex}

\newpage

\appendices

\input{tex/6_Appendix}

\end{document}

%% file: tex/0_Abstract.tex
\IEEEtitleabstractindextext{%
\begin{abstract}
\justifying
We present PyMAF-X, a regression-based approach to recovering parametric full-body models from monocular images. This task is very challenging since minor parametric deviation may lead to noticeable misalignment between the estimated mesh and the input image. Moreover, when integrating part-specific estimations into the full-body model, existing solutions tend to either degrade the alignment or produce unnatural wrist poses. To address these issues, we propose a Pyramidal Mesh Alignment Feedback (PyMAF) loop in our regression network for well-aligned human mesh recovery and extend it as PyMAF-X for the recovery of expressive full-body models. The core idea of PyMAF is to leverage a feature pyramid and rectify the predicted parameters explicitly based on the mesh-image alignment status. Specifically, given the currently predicted parameters, mesh-aligned evidence will be extracted from finer-resolution features accordingly and fed back for parameter rectification. To enhance the alignment perception, an auxiliary dense supervision is employed to provide mesh-image correspondence guidance while spatial alignment attention is introduced to enable the awareness of the global contexts for our network. When extending PyMAF for full-body mesh recovery, an adaptive integration strategy is proposed in PyMAF-X to produce natural wrist poses while maintaining the well-aligned performance of the part-specific estimations. The efficacy of our approach is validated on several benchmark datasets for body, hand, face, and full-body mesh recovery, where PyMAF and PyMAF-X effectively improve the mesh-image alignment and achieve new state-of-the-art results. The project page with code and video results can be found at \href{https://zhanghongwen.cn/pymaf-x}{https://zhanghongwen.cn/pymaf-x}.
\end{abstract}

\begin{IEEEkeywords}
Expressive human mesh recovery, full-body motion capture, monocular 3D reconstruction, mesh alignment feedback.
\end{IEEEkeywords}}

%% file: tex/1_Introduction.tex
\IEEEraisesectionheading{\section{Introduction}\label{sec:introduction}}

\IEEEPARstart{R}ecent years have witnessed the rise of the regression-based paradigm in recovering body~\cite{kanazawa2018end,kolotouros2019convolutional,kolotouros2019learning,moon2020i2l,kocabas2021pare,zhang2021pymaf}, hand~\cite{Zhang_2019_ICCV,zimmermann2019freihand,moon2020interhand2,li2022interacting}, face~\cite{Feng2018,Jackson2017,Sanyal2019_ringnet,DECA_2020}, and even full-body~\cite{choutas2020monocular,rong2020frankmocap,feng2021collaborative,moon2022Hand4Whole} models from monocular images.
These methods~\cite{kanazawa2018end,pavlakos2018learning,kolotouros2019convolutional,kolotouros2019learning} learn to predict model parameters directly from images in a data-driven manner.
Despite the high efficiency and promising results, regression-based methods typically suffer from coarse alignment between the predicted meshes and image observations.

\begin{figure}[t]
	\centering
    \begin{subfigure}[b]{0.35\textwidth}
        \includegraphics[height=35mm]{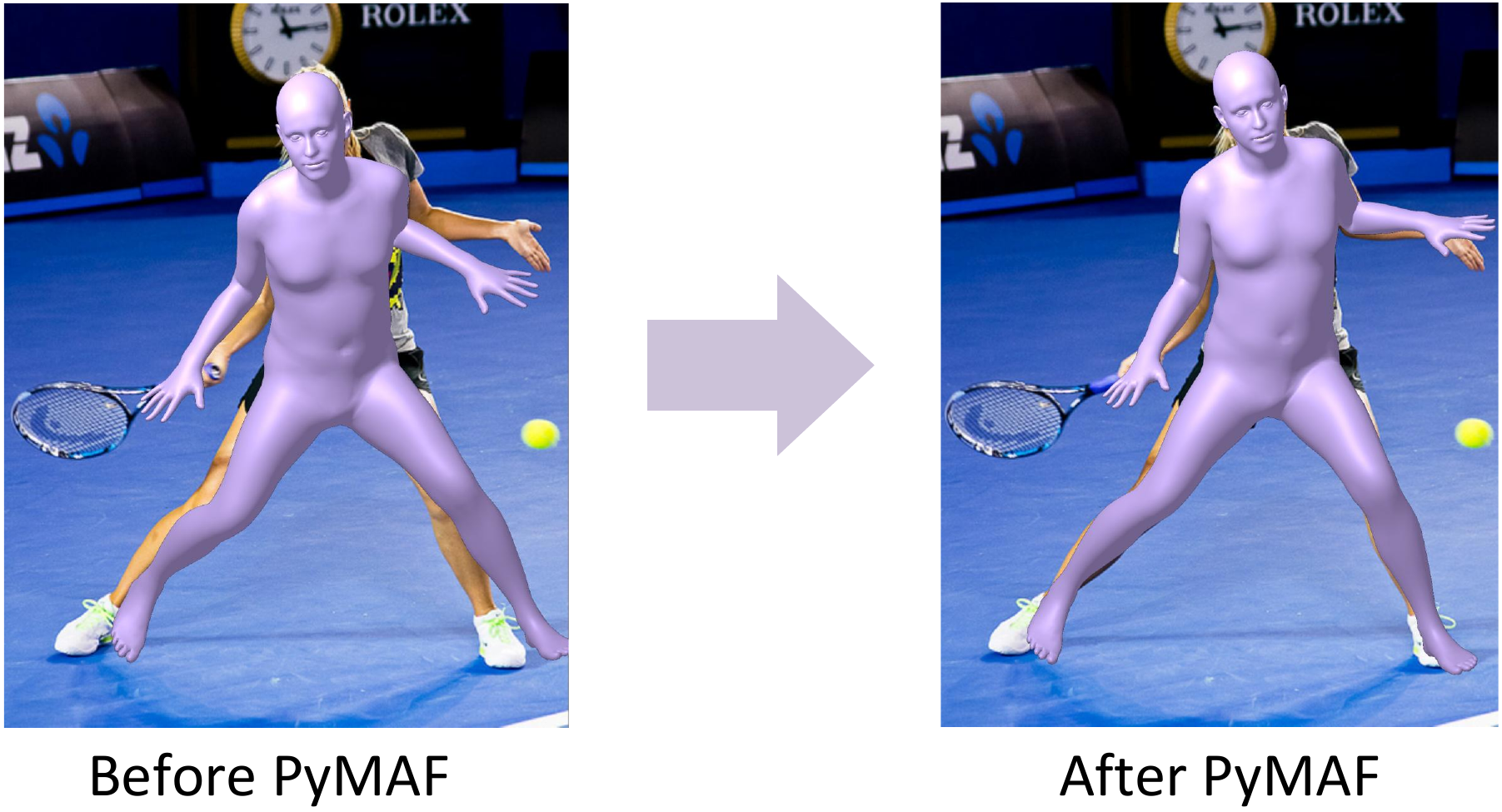}
        \vspace{-3mm}
    \end{subfigure}
    \begin{subfigure}[b]{0.5\textwidth}
        \centering
		\includegraphics[height=31mm]{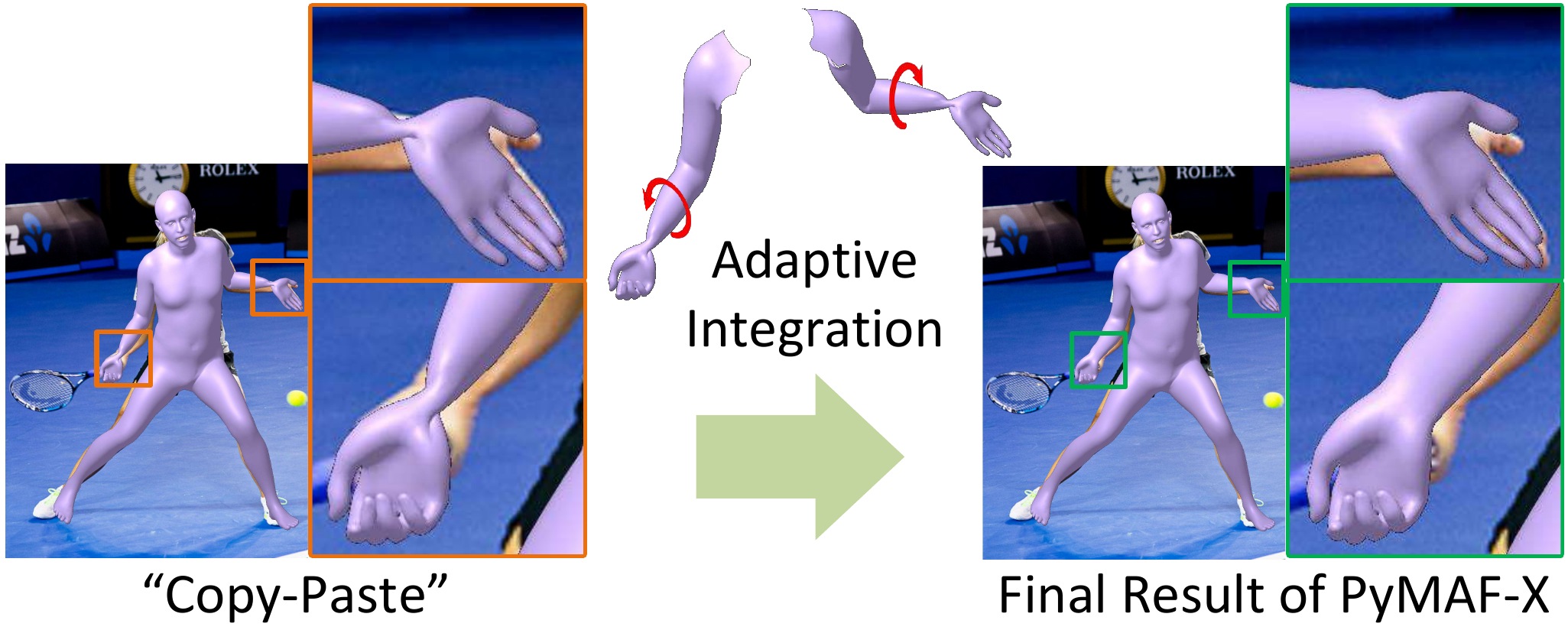}
    \end{subfigure}
    \vspace{-7mm}
	\caption{Top: PyMAF improves the mesh-image alignment of the estimated mesh. Bottom: PyMAF-X produces well-aligned full-body meshes with natural wrist poses.}
	\vspace{-5mm}
	\label{fig:pymafx_teaser}
\end{figure}

\begin{figure}[t]
	\centering
    \begin{subfigure}[b]{0.14\textwidth}
		\includegraphics[height=32mm]{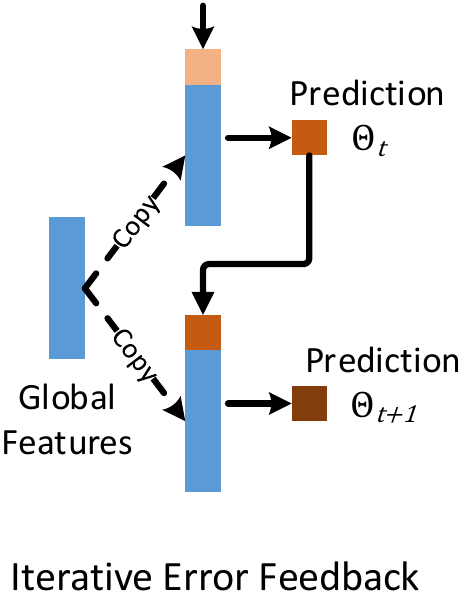}
		\caption{}
		\label{fig:ief}
    \end{subfigure}
    \begin{subfigure}[b]{0.19\textwidth}
		\includegraphics[height=32mm]{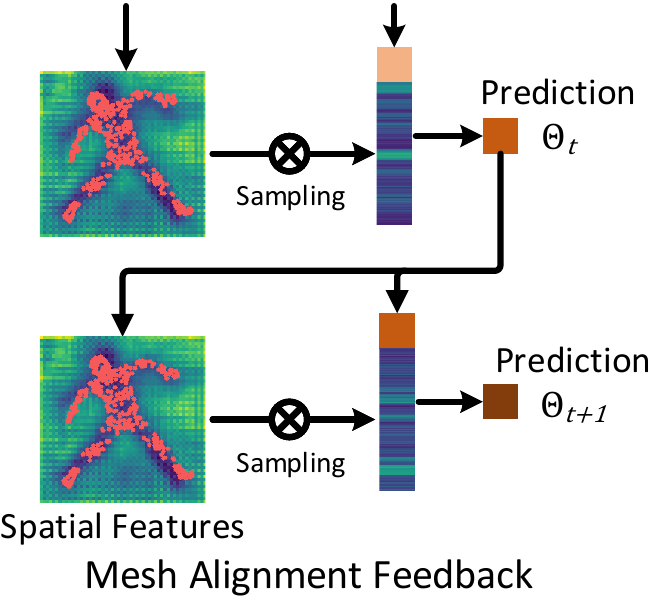}
		\caption{}
		\label{fig:maf}
    \end{subfigure}
    \begin{subfigure}[b]{0.13\textwidth}
		\includegraphics[height=32mm]{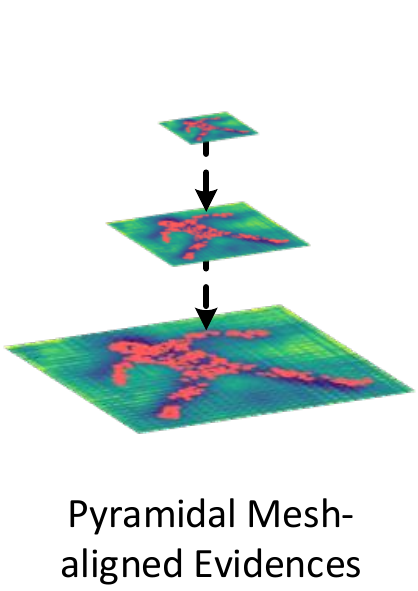}
		\caption{}
		\label{fig:pyramid}
    \end{subfigure}
    \vspace{-3mm}
	\caption{(a) The commonly-used iterative error feedback. (b) The proposed mesh alignment feedback. (c) Mesh-aligned evidence extracted from a feature pyramid.}
	\vspace{-5mm}
	\label{fig:teaser}
\end{figure}

When recovering the parametric body or full-body models~\cite{loper2015smpl,pavlakos2019expressive}, minor rotation errors accumulated along the kinematic chain may lead to noticeable drifts in joint positions (see the top-left example in Fig.~\ref{fig:pymafx_teaser}), since joint poses are represented as relative rotations \wrt their parent joints.
In order to generate well-aligned results, optimization-based methods~\cite{bogo2016keep,lassner2017unite,pavlakos2019expressive} include data terms in the objective function so that the alignment between the projection of meshes and 2D evidence can be optimized explicitly.
Similar strategies are also exploited in regression-based methods~\cite{kanazawa2018end,pavlakos2018learning,kolotouros2019convolutional,kolotouros2019learning} to impose 2D supervisions upon the projection of estimated meshes in the training procedure.
However, during testing, these deep regressors either are open-loop or simply include an Iterative Error Feedback (IEF) loop~\cite{kanazawa2018end} in their architectures.
As shown in Fig.~\ref{fig:ief}, IEF reuses the same global feature in its feedback loop, making the regressor hardly perceive the mesh-image misalignment in the inference phase.

As suggested in previous works~\cite{ronneberger2015u,newell2016stacked,lin2017feature,sun2019deep}, neural networks tend to retain high-level information and discard detailed local features when reducing the spatial size of feature maps.
In order to leverage spatial information in the regression networks, it is essential to extract pixel-wise contexts for fine-grained perception.
Several attempts have been made to leverage pixel-wise representation such as part segmentation~\cite{omran2018neural} or dense correspondences~\cite{xu2019denserac,zhang2020learning} in their regression networks.
Though pixel-level evidence is considered, it is still challenging for those methods to learn structural priors and get hold of spatial details simultaneously based merely on high-resolution contexts.

Motivated by the above observation, we design a Pyramidal Mesh Alignment Feedback (PyMAF) loop in our regression network to exploit multi-scale and position-sensitive contexts for better mesh-image alignment.
The central idea of our approach is to correct parametric deviation explicitly and progressively based on the alignment status.
In PyMAF, mesh-aligned evidence will be extracted from the spatial features according to the 2D projection of the estimated mesh and then fed back to the regressors for parameter updates.
As illustrated in Fig.~\ref{fig:teaser}, the mesh alignment feedback loop takes advantage of more informative features for parameter correction compared with the commonly used iterative error feedback loop~\cite{kanazawa2018end,carreira2016human}.
In order to leverage multi-scale contexts, mesh-aligned evidence is extracted from a feature pyramid so that the coarse-aligned meshes can be corrected with large step sizes based on the lower-resolution features.
To enhance these mesh-aligned features, an auxiliary task is imposed on the highest-resolution feature to infer pixel-wise dense correspondence, guiding the image encoder to preserve the most related information in the spatial feature maps.
Meanwhile, a spatial alignment attention mechanism is introduced to fuse the grid and mesh-aligned features so that the regressor could be aware of the whole image contexts.

Since the SMPL family includes the hand~\cite{romero2017embodied} and face~\cite{li2017learning} models, PyMAF can be easily modified to reconstruct the hand and face meshes.
We leverage three part-specific PyMAF networks as part experts to predict body, hand, and face parameters, and propose PyMAF-X for expressive full-body mesh recovery.
Benefiting from the well-aligned results of each PyMAF-based expert, PyMAF-X can produce plausible full-body mesh results in common scenarios even using the most naive integration strategy~\cite{rong2020frankmocap}.
However, as shown in Fig.~\ref{fig:pymafx_teaser}, the naive ``Copy-Paste'' integration may lead to unnatural wrist poses under challenging cases.
To address this issue, we propose an adaptive integration strategy to adjust the twist rotation of the elbow poses so that the elbow and wrist poses could be more compatible.
In this way, the updated twist rotation of the elbow joint serves as compensation for the wrist joint and helps to produce natural wrist poses in the full-body model.
Moreover, since the twist component~\cite{grassia1998practical} of the elbow poses is the rotation around the elbow-to-wrist bone, it barely changes the position of the body and hand joints, which is the key to maintaining the well-aligned performances of body and hand experts.
Different from existing full-body solutions~\cite{choutas2020monocular,rong2020frankmocap,feng2021collaborative,moon2022Hand4Whole}, our method do not rely on additional networks to infer the wrist poses, and hence bypass the learning issue raised by insufficient full-body mesh annotations.

The contributions of this work can be summarized as follows:
\begin{itemize}[leftmargin=*]
\itemsep0em 
    \item A mesh alignment feedback loop is proposed for regression-based human mesh recovery, where mesh-aligned evidence is exploited to correct parametric errors explicitly so that the estimated meshes can be better aligned with the input images.
    \item A feature pyramid is incorporated with the mesh alignment feedback loop so that the regression network can leverage multi-scale contexts. This yields the Pyramidal Mesh Alignment Feedback (PyMAF) loop, a novel architecture for human mesh recovery.
    \item An auxiliary pixel-wise supervision and spacial alignment attention are introduced in PyMAF to enhance the mesh-aligned features such that they can be more informative, relevant, and aware of the whole image contexts.
    \item PyMAF is further extended as PyMAF-X for full-body mesh recovery, where an adaptive integration strategy with the elbow-twist compensation is proposed to avoid unnatural wrist poses while maintaining the alignment of the body and hand estimations.
\end{itemize}

An early version of this work has been published as a conference paper~\cite{zhang2021pymaf}. We have made significant extensions to our previous work~\cite{zhang2021pymaf} from three aspects.
First, PyMAF is improved to be more accurate with the newly introduced spatial alignment attention, which effectively enhances the feature learning and further improves the mesh-image alignment.
Second, PyMAF goes beyond body mesh recovery and is extended to reconstruct hand and full-body models from monocular images. The well-aligned performance of the body- and hand-specific PyMAF makes it more promising to produce well-aligned full-body meshes.
Third, an adaptive integration strategy is proposed to assemble predictions from body and hand experts.
Such a strategy effectively addresses the unnatural wrist issues while maintaining the part-specific alignment.
Based on these updates, our final method PyMAF-X achieves new state-of-the-art results both qualitatively and quantitatively, contributing novel solutions towards the well-aligned and natural recovery of full-body models from monocular images.

%% file: tex/2_RelatedWork.tex
\section{Related Work}\label{sec:related_work}

\subsection{Monocular Human Mesh Recovery}
Monocular recovery of human meshes has been actively studied in recent years. Aiming at the same goal of producing well-aligned and natural results, two different paradigms for human mesh recovery have been investigated in the research community. In this subsection, we give a brief review of these two paradigms and refer readers to~\cite{tian2022recovering} for a more comprehensive survey.

\textbf{Optimization-based Approaches.}
Pioneering work in this field mainly focus on the optimization process of fitting parametric models (\eg, SCAPE~\cite{anguelov2005scape} and SMPL~\cite{loper2015smpl}) to 2D observations such as keypoints and silhouettes~\cite{sigal2008combined,guan2009estimating,bogo2016keep}.
In their objective functions, prior terms are designed to penalize the unnatural shape and pose, while data terms measure the fitting errors between the re-projection of meshes and 2D evidence.
Based on this paradigm, different updates have been investigated to incorporate information such as 2D/3D body joints~\cite{bogo2016keep,zhang2021lightweight}, silhouettes~\cite{lassner2017unite,huang2017towards}, part segmentation~\cite{zanfir2018monocular}, dense correspondences~\cite{guler2019holopose} in the fitting procedure.
Despite the well-aligned results obtained by these optimization-based methods, their fitting process tends to be slow and sensitive to initialization.
Recently, Song~\etal~\cite{song2020human} exploit the learned gradient descent in the fitting process.
Though this solution leverages rich 2D pose datasets and alleviates many issues in traditional optimization-based methods, it still relies on the accuracy of 2D poses and breaks the end-to-end learning.
Alternatively, our solution supports end-to-end learning and is also able to leverage rich 2D datasets thanks to the progress (\eg, SPIN~\cite{kolotouros2019learning}, EFT~\cite{joo2021exemplar}, and NeuralAnnot~\cite{moon2020neuralannot}) in the generation of more precise pseudo 3D ground-truth for 2D datasets~\cite{andriluka20142d,johnson2010clustered,lin2014microsoft}.

\textbf{Regression-based Approaches.}
Alternatively, taking advantage of the powerful nonlinear mapping capability of neural networks, recent regression-based approaches~\cite{kanazawa2018end,pavlakos2018learning,omran2018neural,kolotouros2019learning,choutas2020monocular,choi2020pose2mesh,li2021hybrik,lin2021end} have made significant advances in predicting human models directly from monocular images.
These deep regressors take 2D evidence as input and learn model priors implicitly in a data-driven manner under different types of supervision signals~\cite{tung2017self,kanazawa2018end,pavlakos2019texturepose,rong2019delving,doersch2019sim2real,zanfir2020weakly,kundu2020appearance,dwivedi2021learning} during the learning procedure.
To mitigate the learning difficulty of the regressor, different network architectures have also been designed to leverage proxy representations such as silhouette~\cite{pavlakos2018learning,varol2018bodynet}, 2D/3D joints~\cite{tung2017self,guler2019holopose,pavlakos2018learning,moon2020i2l,choi2020pose2mesh,zanfir2021thundr,choi20213dcrowdnet,li2021hybrik,moon2022Hand4Whole}, segmentation~\cite{omran2018neural,rueegg2020chained} and dense correspondences~\cite{xu2019denserac,zhang2020learning}.
Such strategies can benefit from synthetic data~\cite{xu2019denserac,sengupta2020synthetic} and the progress in the estimation of proxy representations~\cite{pavlakos2017coarse,cao2019openpose,guler2018densepose,sun2019deep,wu2021graph}.
In these regressors, though supervision signals are imposed on the re-projected models to penalize the mismatched predictions during training, their architectures can hardly perceive the misalignment during the inference phase.
In comparison, the proposed PyMAF is a close-loop for both training and inference, which enables a feedback loop in our regressor to leverage spatial evidence for better mesh-image alignment of the estimated human models.

Directly regressing model parameters from images is very challenging, even for neural networks.
Existing methods have also offered non-parametric solutions to reconstruct human body models.
Among them, volumetric representation~\cite{varol2018bodynet,zheng2019deephuman}, mesh vertices~\cite{kolotouros2019convolutional,lin2021end,lin2021mesh}, and position maps~\cite{yao2019densebody,zhang2020object,zeng20203d,wang2022best} have been adopted as regression targets.
Using non-parametric representations as the regression targets is more readily to leverage high-resolution features but needs further processing to retrieve parametric models from the outputs.
Besides, the mesh surfaces of non-parametric outputs tend to be rough and more sensitive to occlusions without additional structure priors.
In our solution, the deep regressor uses spatial features at multiple scales for both high-level and fine-grained perception. It produces parametric models directly with no further processing required.

Recently, there are also numerous efforts devoted to achieving or handling multi-person recovery~\cite{fieraru2020three,jiang2020coherent,zhang2021body,sun2021monocular,sun2021putting,choi20213dcrowdnet,fieraru2021remips}, video inputs~\cite{kocabas2020vibe,choi2021beyond,wan2021encoder,zhang2021learning,rempe2021humor,pavlakos2020human}, occlusions~\cite{zhang2020learning, kocabas2021pare,khirodkar2022occluded,yuan2021glamr}, more accurate shape~\cite{sengupta2020synthetic,sengupta2021hierarchical,choutas2022accurate}, ambiguities~\cite{sengupta2021probabilistic,kolotouros2021probabilistic}, camera estimation~\cite{kissos2020beyond,kocabas2021spec}, imbalanced data~\cite{guan2021bilevel,rong2022chasing}, pseudo ground-truth generation~\cite{kolotouros2019learning,joo2021exemplar,moon2020neuralannot}, and clothed human reconstruction~\cite{saito2019pifu,zheng2021pamir,zheng2021deepmulticap,xiu2022icon}.
Our work is complementary to them and focuses on the design of regressor architectures for single-image well-aligned body and full-body mesh recovery.

\begin{figure*}[t]
	\begin{center}
		\includegraphics[width=0.9\textwidth]{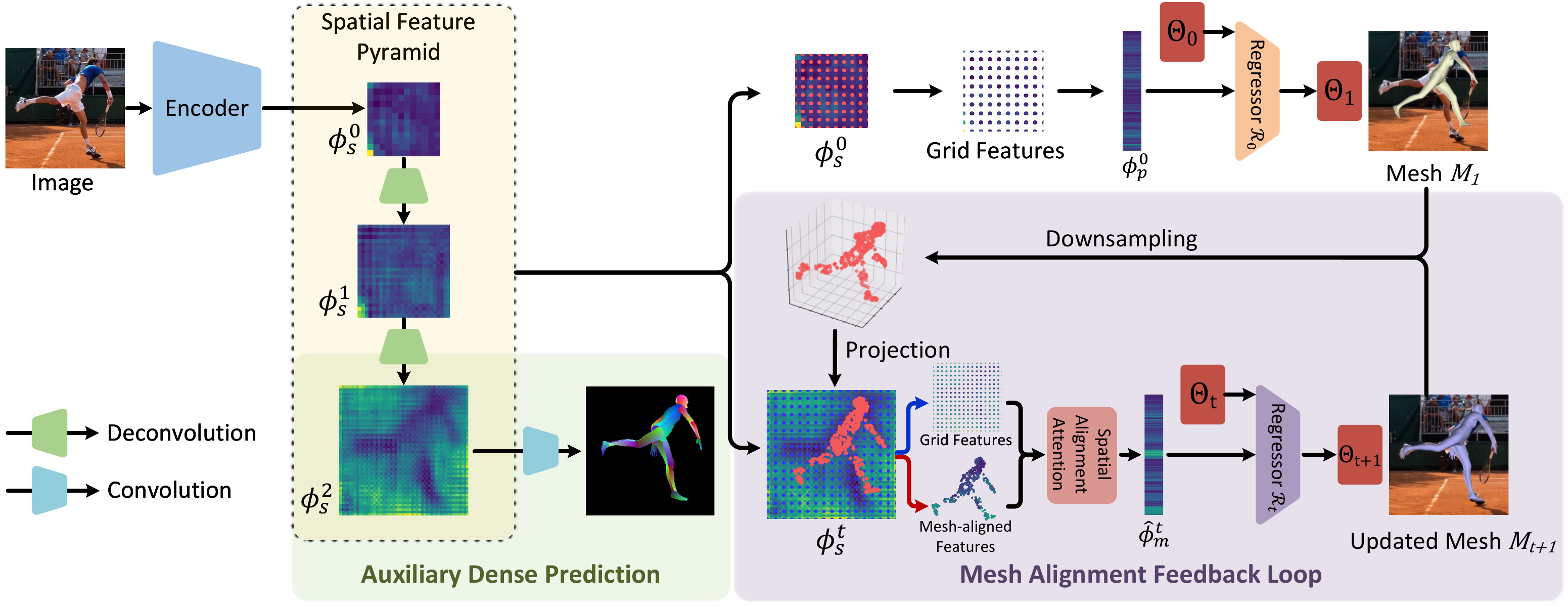}
        \vspace{-3mm}
		\caption{Illustration of the proposed Pyramidal Mesh Alignment Feedback (PyMAF) for human mesh recovery. PyMAF leverages a feature pyramid and enables an alignment feedback loop in our network. Given a coarse-aligned model prediction, mesh-aligned evidence is extracted from finer-resolution features accordingly and fed back to a regressor for parameter rectification.}
		\vspace{-5mm}
		\label{fig:framework}
	\end{center}
\end{figure*}

\subsection{Full-body Mesh Recovery}
Compared with the large number of solutions for the body-only~\cite{kanazawa2018end,kolotouros2019convolutional,kolotouros2019learning,moon2020i2l,zhang2020learning}, hand-only~\cite{Baek_2019_CVPR,Boukhayma2019,hasson_2019_cvpr,kulon2019rec,Zhang_2019_ICCV,Ge2019,Kulon_2020_CVPR,moon2020interhand2,zhang2021interacting,rong2021monocular,li2022interacting}, and face-only~\cite{Feng2018,Jackson2017,Tewari2018,Sanyal2019_ringnet,DECA_2020,wang2022faceverse} mesh recovery, the full-body mesh recovery receives less attention due to its challenging nature and the lack of annotated datasets.
Similar to the developments of body-only mesh recovery algorithms, the research in the field of full-body mesh recovery begins with the proposal of full-body models, including Frank~\cite{joo2018total}, Adam~\cite{joo2018total}, SMPL-X~\cite{pavlakos2019expressive}, and GHUM~\cite{xu2020ghum}, \etc, and their corresponding optimization-based methods~\cite{joo2018total,xiang2019monocular,pavlakos2019expressive,xu2020ghum,li2020full,zhang2021lightweight}.
Recently, several regression-based methods~\cite{choutas2020monocular,rong2020frankmocap,sun2022learning,feng2021collaborative,moon2022Hand4Whole} have been proposed to overcome the slow and unnatural issues of optimization-based methods.

Following the pioneering work ExPose~\cite{choutas2020monocular}, regression-based methods~\cite{choutas2020monocular,rong2020frankmocap,zhou2021monocular,zanfir2021neural,feng2021collaborative,moon2022Hand4Whole,grishchenko2022blazepose} typically consist of three part-specific modules, namely part experts, to predict parameters of body, hand, and face from the corresponding part images cropped from original inputs.
They differ mainly in the architecture of the part experts and the strategy to integrate part estimations.
As the part experts are basically chosen from the body- or hand-only mesh recovery solutions, the integration strategy to sew up independent estimations becomes an essential aspect of a regression-based full-body method.
The most straightforward strategy to integrate the body and hand estimations would be the ``Copy-Paste''~\cite{choutas2020monocular,rong2020frankmocap}.
To obtain more natural integration results, learning-based strategies are proposed in recent state-of-the-art methods~\cite{rong2020frankmocap,zhou2021monocular,feng2021collaborative}.
For instance, FrankMocap~\cite{rong2020frankmocap} learns to correct the arm poses based on the distance between the wrist positions predicted by body and hand experts.
Zhou~\etal~\cite{zhou2021monocular} incorporate body features in the learning of the hand expert so that the predicted hand poses could be more compatible with the arm.
PIXIE~\cite{feng2021collaborative} introduces a learnable moderator to merge body and hand features for the regression of wrist and finger poses.
All the above solutions rely on additional networks to predict or correct the wrist poses with the condition of body information, which is typically inferior to the original hand poses predicted by the hand expert, resulting in degraded alignment on the hand parts.
Recently, Hand4Whole~\cite{moon2022Hand4Whole} proposes to learn wrist poses based on the positions of selected hand joints but does not consider the compatibility of arm poses.
In contrast to existing solutions, PyMAF-X resorts to the adjustment of the twist components~\cite{grassia1998practical} of wrist and elbow poses, which produces natural wrist rotations while maintaining the well-aligned performances of each part expert during the integration.
Besides, our motivation and method also differ from the previous work~\cite{nakatsuka2021learning,li2021hybrik} that decomposes the twist components in the inverse kinematics problem.

\subsection{Iterative Fitting in Regression Tasks}
Strategies for incorporating fitting processes along with regression tasks have also been investigated in the literature.
For human mesh recovery, Kolotouros~\etal~\cite{kolotouros2019learning} combine an iterative fitting procedure with the training procedure to generate more accurate ground truths for better supervision.
Several attempts have been made to deform human meshes so that they can be aligned with the intermediate estimates such as depth maps~\cite{zhu2019detailed}, part segmentation~\cite{zanfir2021neural}, and dense correspondences~\cite{guler2019holopose}.
These approaches adopt intermediate estimations as fitting objectives and hence rely on their quality.
In contrast, our approach uses the currently estimated meshes to extract deep features for refinement, enabling the fully end-to-end learning of the deep regressor.

In a broader view, remarkable efforts have been made to involve iterative fitting strategies in other computer vision tasks, including facial landmark localization~\cite{xiong2013supervised,trigeorgis2016mnemonic,chandran2020attention}, human/hand pose estimation~\cite{oberweger2019generalized,carreira2016human}, \etc.
For generic objects, Pixel2Mesh~\cite{wang2018pixel2mesh} progressively deforms an initial ellipsoid by leveraging perceptual features.
Following the spirit of these works, we exploit new strategies to extract fine-grained evidence and contribute novel solutions in the context of human mesh recovery.

%% file: tex/3_Methodology.tex
\section{Method}\label{sec:methodology}

In this section, we will elaborate technical details of our approach.
We first present PyMAF, a powerful model for regression-based human mesh recovery, then extend it to PyMAF-X for full-body mesh recovery.

\subsection{PyMAF for Body-only Mesh Recovery}
As illustrated in Fig.~\ref{fig:framework}, PyMAF consists of a feature pyramid for mesh recovery in a coarse-to-fine fashion.
Coarse-aligned predictions will be improved by utilizing the mesh-aligned evidence extracted from spatial feature maps.
In order to enhance the mesh-aligned evidence, an auxiliary dense prediction task is imposed on the image encoder while a spatial alignment attention is applied to fuse the grid and mesh-aligned features.

\subsubsection{Feature Pyramid for Body Model Regression}
Our image encoder aims to generate a pyramid of spatial features from coarse to fine granularities, which provide descriptions of the posed person at different scale levels.
The feature pyramid will be used in subsequent predictions of the SMPL model with the pose, shape, and camera parameters $\Theta=\{\bm{\theta}, \bm{\beta}, \bm{\pi}\}$.

Formally, the encoder takes an image $I$ as input and outputs a set of spatial features $\{\bm{\phi}_s^t \in \mathbb{R}^{C_s\times H_s^t \times W_s^t}\}_{t=0}^{T-1}$ at the end, where $H_s^t$ and $W_s^t$ are monotonically increasing.
At level $t$, based on the feature map $\bm{\phi}_s^t$, a set of sampling points $X^t$ will be used to extract point-wise features.
Specifically, for each 2D point $x$ in $X^t$, point-wise features $\bm{\phi}_s^t(x) \in \mathbb{R}^{C_s\times 1}$ will be extracted from $\bm{\phi}_s^t$ accordingly using the bilinear sampling.
These point-wise features will go through a MLP (multi-layer perceptron) for dimension reduction and be further concatenated together as a feature vector $\bm{\phi}_p^t$, \ie,
\begin{equation}
    \bm{\phi}_p^t = \mathcal{F}(\bm{\phi}_s^t, X^t) = \oplus\left(\left\{f\left(\bm{\phi}_s^t(x)\right), \text{for}~x~\text{in}~X^t\right\}\right),
\label{ptfeat}
\end{equation}
where $\mathcal{F}(\cdot)$ denotes the feature sampling and processing operations, $\oplus$ denotes the concatenation, and $f(\cdot)$ is the MLP.
After that, a parameter regressor $\mathcal{R}_t$ takes features $\bm{\phi}_p^t$ and the current estimation of parameters $\Theta_t$ as inputs and outputs the parameter residual.
Parameters are then updated as $\Theta_{t+1}$ by adding the residual to $\Theta_t$.
For the level $t=0$, $\Theta_0$ adopts the mean parameters calculated from training data.

Given the parameter predictions $\Theta$ (the subscript $t$ is omitted for simplicity) at each level, a mesh with vertices of $M=\mathcal{M}(\bm{\theta}, \bm{\beta}) \in \mathbb{R}^{N\times 3}$ can be generated accordingly, where $N = 6890$ denotes the number of vertices in the SMPL model.
These mesh vertices are mapped to sparse 3D joints $J\in \mathbb{R}^{N_j\times 3}$ by a pretrained linear regressor, and further projected on the image coordinate system as 2D keypoints $K=\bm{\Pi}(J)\in \mathbb{R}^{N_j\times 2}$, where $\bm{\Pi}(\cdot)$ denotes the projection function based on the camera parameters $\bm{\pi}$.
Note that the pose parameters in $\Theta$ are represented as relative rotations along kinematic chains, and minor parameter errors can lead to noticeable misalignment between the 2D projection and image evidence.
To penalize such misalignment during the training of the regression network, we follow common practices~\cite{kanazawa2018end,kolotouros2019learning} to add 2D supervisions on the 2D keypoints projected from the estimated mesh.
Meanwhile, additional 3D supervisions on 3D joints and model parameters are added when ground truth 3D labels are available.
Overall, the loss function for the parameter regressor is written as
\begin{equation}
\mathcal{L}_{reg} = \lambda_{2d}||K - \hat{K}|| + \lambda_{3d}||J - \hat{J}|| + \lambda_{para}||\Theta - \hat{\Theta}||,
\label{eq:loss_reg}
\end{equation}
where $||\cdot||$ is the squared L2 norm, $\hat{K}$, $\hat{J}$, and $\hat{\Theta}$ denote the ground truth 2D keypoints, 3D joints, and model parameters, respectively.

One of the improvements over the commonly used parameter regressors is that our regressors can better leverage spatial information.
Unlike the commonly used regressors taking the global features $\bm{\phi}_g \in \mathbb{R}^{C_g\times 1}$ as input, our regressor uses the point-wise information obtained from spatial features $\bm{\phi}_s^t$.
A straightforward strategy to extract point-wise features would be using grid-pattern points $X_{grid}$ and uniformly sampling features from $\bm{\phi}_s^t$.
In the proposed approach, the sampling points $X^t$ adopt the grid pattern at the level $t=0$ and will be updated according to the current estimates when $t>0$.
We will show that such a mesh conditioned sampling strategy helps the regressor produce well-aligned reconstruction results.

\subsubsection{Mesh Alignment Feedback Loop}

As mentioned in HMR~\cite{kanazawa2018end}, directly regressing mesh parameters in one go is challenging.
To tackle this issue, HMR uses an Iterative Error Feedback (IEF) loop to iteratively update $\Theta$ by taking the global features $\bm{\phi}_g$ and the current estimation of $\Theta$ as input.
Though the IEF strategy reduces parameter errors progressively, it uses the same global features each time for parameter update, which lacks fine-grained information and is not adaptive to new predictions.
By contrast, we propose a Mesh Alignment Feedback (MAF) loop so that mesh-aligned evidence can be leveraged in our regressor to rectify current parameters and improve the mesh-image alignment of the estimated model.

\paragraph{Mesh-aligned Features.}
In the proposed mesh alignment feedback loop, we extract mesh-aligned features from $\bm{\phi}_s^t$ based on the currently estimated mesh $M_t$ when $t>0$ to obtain more fine-grained and position-sensitive evidence.
Compared with the global features or the uniformly sampled grid features, mesh-aligned features can reflect the mesh-image alignment status of the current estimation, which is more informative for parameter rectification.
Specifically, the sampling points $X^t$ are set as the mesh-aligned points $X_{mesh}^t$, which are obtained by first down-sampling the mesh $M_t$ to $\tilde{M}_t$ and then projecting it on the 2D image plane, \ie, $X^t = X_{mesh}^t = \bm{\Pi}(\tilde{M}_t)$.
Based on $X_{mesh}^t$, the mesh-aligned features $\bm{\phi}_m^t$ will be extracted from $\bm{\phi}_s^t$ using Eq.~\ref{ptfeat}, \ie, 
\begin{equation}
\bm{\phi}_m^t = \bm{\phi}_p^t = \mathcal{F}(\bm{\phi}_s^t, \bm{\Pi}(\tilde{M}_t)).
\end{equation}

\begin{figure}[t]
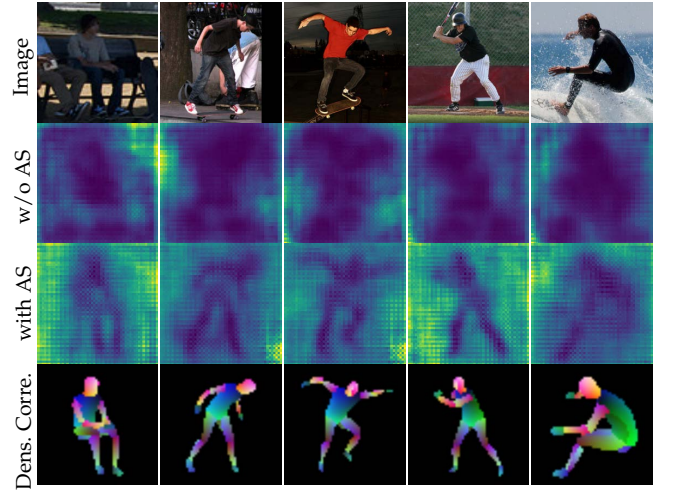

	\centering
	\begin{tikzpicture}[remember picture,overlay]
	\node[font=\fontsize{8pt}{8pt}\selectfont, rotate=90] at (-4.2,-0.85) {Image};
	\node[font=\fontsize{8pt}{8pt}\selectfont, rotate=90] at (-4.2,-2.45) {w/o AS};
	\node[font=\fontsize{8pt}{8pt}\selectfont, rotate=90] at (-4.2,-4.) {with AS};
	\node[font=\fontsize{8pt}{8pt}\selectfont, rotate=90] at (-4.2,-5.7) {Dens. Corre.};
	\end{tikzpicture}
    \\
	\foreach \idx in {1,2,3,4,5} {
    	\begin{subfigure}[h]{0.08\textwidth}
    	    \includegraphics[width=1.1\textwidth]{fig/vis/fm/img_\idx.pdf}
        \end{subfigure}
	}
	\\
	\foreach \idx in {1,2,3,4,5} {
    	\begin{subfigure}[h]{0.08\textwidth}
    	    \includegraphics[width=1.1\textwidth]{fig/vis/fm/wo_dp_\idx.pdf}
        \end{subfigure}
	}
	\\
	\foreach \idx in {1,2,3,4,5} {
    	\begin{subfigure}[h]{0.08\textwidth}
    	    \includegraphics[width=1.1\textwidth]{fig/vis/fm/dp_\idx.pdf}
        \end{subfigure}
	}
	\\
	\foreach \idx in {1,2,3,4,5} {
    	\begin{subfigure}[h]{0.08\textwidth}
    	    \includegraphics[width=1.1\textwidth]{fig/vis/fm/iuv_\idx.pdf}
        \end{subfigure}
	}
	\\
	\caption{Visualization of the spatial feature maps and predicted dense correspondences. Top: Input images. Second / Third Row: Spatial feature maps learned without/with Auxiliary Supervision (AS). Bottom: Predicted dense correspondence maps under auxiliary supervision.}
	\vspace{-3mm}
	\label{fig:fm_auxsupv}
\end{figure}

\begin{figure*}[t]
	\begin{center}
		\includegraphics[width=0.8\textwidth]{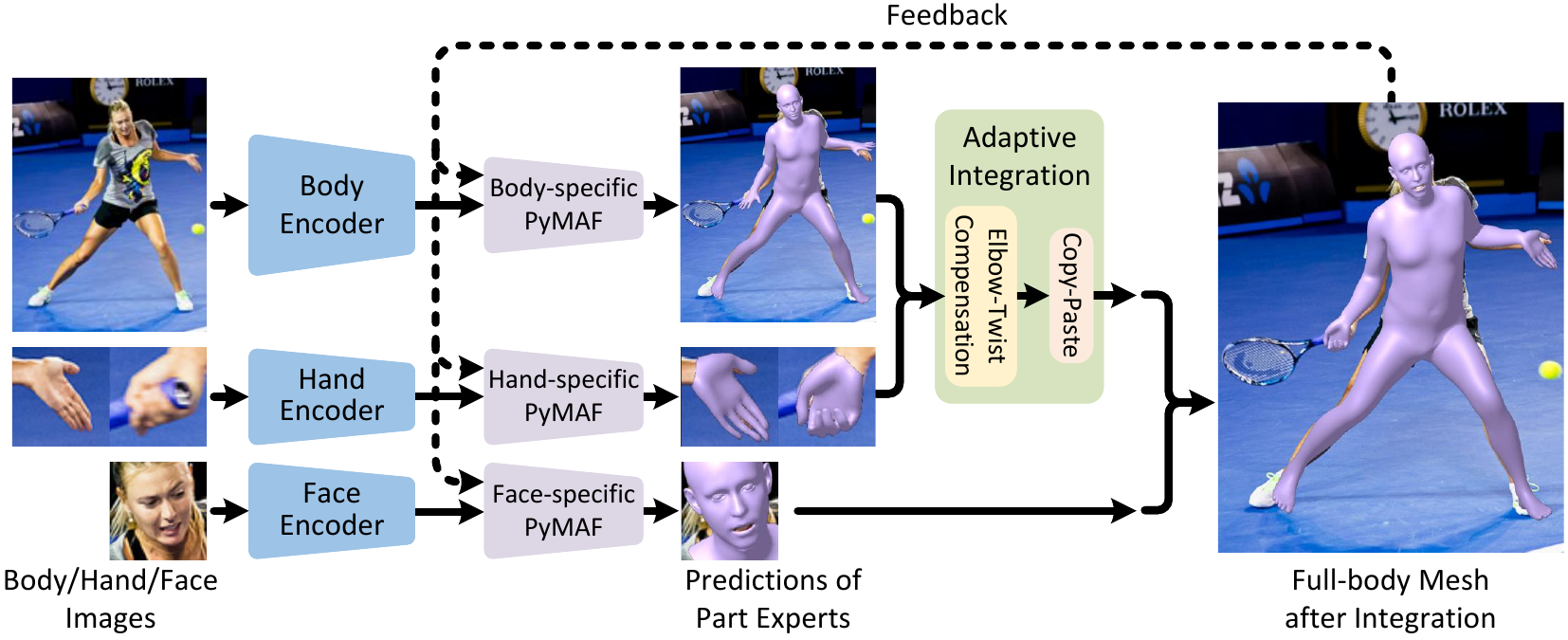}
		\vspace{-3mm}
		\caption{The overall pipeline of PyMAF-X for full-body mesh recovery. PyMAF-X consists of three part-specific PyMAFs for part mesh prediction and integrates them together via the proposed adaptive integration strategy.}
		\vspace{-5mm}
		\label{fig:pymaf-x}
	\end{center}
\end{figure*}

\paragraph{Spatial Alignment Attention.}
Though the mesh-aligned features $\bm{\phi}_m^t$ are position-sensitive, these features are confined to the re-projection regions of the current mesh result. To enable the perception of the relative positions in the whole image context, we further design spatial alignment attention to fuse the information from both grid and mesh-aligned features.
Considering that both these two features are extracted from the same spatial feature map, we adopt a self-attention module to process them.
Specifically, the point-wise features extracted based on the grid-pattern points $X_{grid}$ and the mesh-aligned points $X_{mesh}^t$ are first concatenated together as $\bm{\phi}_{gm}^t$:
\begin{equation}
\bm{\phi}_{gm}^t = \left\{\bm{\phi}_s^t(x), \text{for}~x~\text{in}~\{X_{grid}\bigcup X_{mesh}^t\}\right\} \in \mathbb{R}^{N_{gm}\times C_s},
\end{equation}
where $N_{gm}$ is the total number of the grid-pattern and mesh-aligned points.
Then, spatial alignment attention is applied to learn attentive relations among $\bm{\phi}_{gm}^t$ so that the mesh-aligned features can be more effectively enhanced with the spatial information in the grid features.
In our solution, a self-attention module~\cite{vaswani2017attention} is employed to process the features $\bm{\phi}_{gm}^t$:
\begin{equation}
\begin{aligned}
\bm{Q,K,V} &= \bm{\phi}_{gm}^t\bm{W}^{Q},\bm{\phi}_{gm}^t\bm{W}^{K},\bm{\phi}_{gm}^t\bm{W}^{V},\\
\bm{\hat{\phi}}_{gm}^t &= Att(\bm{Q,K})\bm{V},
\end{aligned}
\end{equation}
where $\bm{W}^{Q}$, $\bm{W}^{K}$, and $\bm{W}^{V}$ are the learnable matrices used to generate different subspace representations of the query, key, and value features $\bm{Q,K,V}$, respectively, $Att(\cdot)$ denotes the scaled dot-product attention function~\cite{vaswani2017attention} with softmax.
In this way, the messages of the grid and mesh-aligned features can be fully fused together since the self-attention mechanism captures the relationships between all elements of the features $\bm{\phi}_{gm}^t$.
After that, the enhanced mesh-aligned features $\bm{\hat{\phi}}_{m}^t$ are obtained by reducing the dimension of $\bm{\hat{\phi}}_{gm}^t$ and concatenating them together.
Finally, the enhanced mesh-aligned features $\bm{\hat{\phi}}_{m}^t$ are fed into the regressor $\mathcal{R}_t$ for parameter update:
\begin{equation}
    \Theta_{t+1} = \Theta_t + \mathcal{R}_t\left(\Theta_t, \bm{\hat{\phi}}_{m}^t\right), \text{for~} t > 0.
\end{equation}

\subsubsection{Auxiliary Dense Supervision}

As depicted in the second row of Fig.~\ref{fig:fm_auxsupv}, spatial features tend to be affected by noisy inputs, since raw images may contain a large amount of unrelated information such as occlusions, appearance, and illumination variations.
To improve the reliability of the mesh-aligned cues extracted from spatial features, we impose an auxiliary pixel-wise prediction task on the spatial features at the last level.
Specifically, during training, the spatial feature maps $\bm{\phi}_s^{T-1}$ will go through a convolutional layer to generate dense correspondence maps with pixel-wise supervision.
Dense correspondences encode the mapping relationship between foreground pixels on the 2D image plane and mesh vertices in 3D space. In this way, the auxiliary supervision provides mesh-image correspondence guidance for the image encoder to preserve the most related information in the spatial feature maps.

In our implementation, we adopt the IUV maps defined in DensePose~\cite{guler2018densepose} as the dense correspondence representation, which consists of the part index and UV values of the mesh vertices.
Note that we do not use DensePose annotations in the dataset but render IUV maps based on the ground-truth SMPL models~\cite{zhang2020learning}.
During training, classification and regression losses are applied on the part index $P$ and $UV$ channels of dense correspondence maps, respectively.
Specifically, for the part index $P$ channels, a cross-entropy loss is applied to classify a pixel belonging to either background or one among body parts.
For the $UV$ channels, a smooth L1 loss is applied to regress the corresponding $UV$ values of the foreground pixels.
Only the foreground regions are taken into account in the $UV$ regression loss, \ie, the estimated $UV$ channels are firstly masked by the ground-truth part index channels before applying the regression loss.
Overall, the loss function for the auxiliary pixel-wise supervision is written as
\begin{equation}
\begin{aligned}
\mathcal{L}_{aux} = &\lambda_{pi}CrossEntropy(P, \hat{P}) \\
            + &\lambda_{uv}SmoothL1(\hat{P} \odot U, \hat{P} \odot \hat{U}) \\
            + &\lambda_{uv}SmoothL1(\hat{P} \odot V, \hat{P} \odot \hat{V}),
\end{aligned}
\label{eq:aux_supv}
\end{equation}
where $\odot$ denotes the mask operation.
Note that the auxiliary prediction is required in the training phase only.

Fig.~\ref{fig:fm_auxsupv} visualizes the spatial features of the encoder trained with and without auxiliary supervision, where the feature maps are simply added along the channel dimension as grayscale images and visualized with colormap.
We can see that the spatial features are more neat and robust to input variations after applying auxiliary supervision.
Note that the dense correspondence is not limited to the IUV representation, the Projected Normalized Coordinate Code (PNCC)~\cite{zhu2017face} can be also adopted as dense correspondences when IUV is not defined in the mesh model. More discussions about the choice of dense correspondences can be found in the Supplementary Material.

\subsection{PyMAF-X for Full-body Mesh Recovery}

The body-specific PyMAF can be easily modified to reconstruct hand and face meshes by simply changing the SMPL model in the above formulation to the MANO~\cite{romero2017embodied} and FLAME~\cite{li2017learning} models.
Based on the regression power of PyMAF, we extend it to PyMAF-X for full-body mesh recovery.

Following previous works~\cite{choutas2020monocular,rong2020frankmocap,feng2021collaborative,moon2022Hand4Whole}, PyMAF-X consists of three experts, \ie, three part-specific PyMAFs, to predict the parameters of body, hand, and face, as illustrated in Fig.~\ref{fig:pymaf-x}.
To ensure high-resolution observations of part regions, part experts perform individual predictions on the body, hand, and face images cropped from the original inputs.
At each iteration of the mesh alignment feedback loop, the predictions of the body-, hand-, and face-specific PyMAF are collected and integrated as the parameters $\Theta_{fb}=\{\bm{\theta}_{fb}, \bm{\beta}_{fb}, \bm{\psi}, \bm{\pi}\}$ of the full-body model SMPL-X~\cite{pavlakos2019expressive}, where $\bm{\theta}_{fb}$, $\bm{\beta}_{fb}$, and $\bm{\psi}$ denotes the pose, shape, and facial expression parameters, respectively.
The pose parameters $\bm{\theta}_{fb}$ consist of the rotational poses of 55 joints in total, including 22 joints for the body, 30 finger joints for the hands, and 3 jaw joints for the face.
The camera parameters $\bm{\pi}$ are taken from the predictions of the body-specific PyMAF and used to project body, hand, and face vertices on the image plane.
Moreover, considering that the positions of hand and face are susceptible to inaccurate body pose estimations, we align the center of their re-projected points to the image center of hand and face to ensure their mesh-aligned features are meaningful.

\paragraph{Naive Integration.} After individual regression of each part, we need to figure out the rotation of wrist joints to integrate the body and hand meshes. The most straightforward strategy would be the naive ``Copy-Paste'' integration~\cite{rong2020frankmocap}.
Specifically, the poses of the wrist joints are calculated based on the body poses predicted by the body expert and the global orientation of hands predicted by the hand expert.
Let $\acute{\bm{\theta}}_{hand}$ be the global orientation of the left or right hand, which is also the global rotation of the wrist joint. 
The wrist pose of the full-body model can be solved by first computing the global rotation $\acute{\bm{\theta}}_{elbow}$ of the elbow joint and then the relative rotation $\bm{\theta}_{wrist}$ of the wrist joint, \ie,
\begin{equation}
\begin{aligned}
\acute{\bm{\theta}}_{elbow} &= \prod_{j \in A(elbow)}\bm{\theta}_{j},\\
\bm{\theta}_{wrist} &= \acute{\bm{\theta}}_{hand}^{-1}\acute{\bm{\theta}}_{elbow},
\end{aligned}
\label{eq:wrist}
\end{equation}
where $\bm{\theta}_{j}$ denotes the relative rotation of the $j$-th body joint, $A(elbow)$ the ordered set of joint ancestors of the elbow joint and itself in the kinematic tree, and $\acute{\bm{\theta}}_{hand}^{-1}$ the inverse global rotation of the hand.
Benefiting from the well-aligned results of each part, PyMAF-X can produce plausible results in common scenarios using such a simple integration strategy.

\begin{figure}[t]
	\begin{center}
		\includegraphics[width=0.48\textwidth]{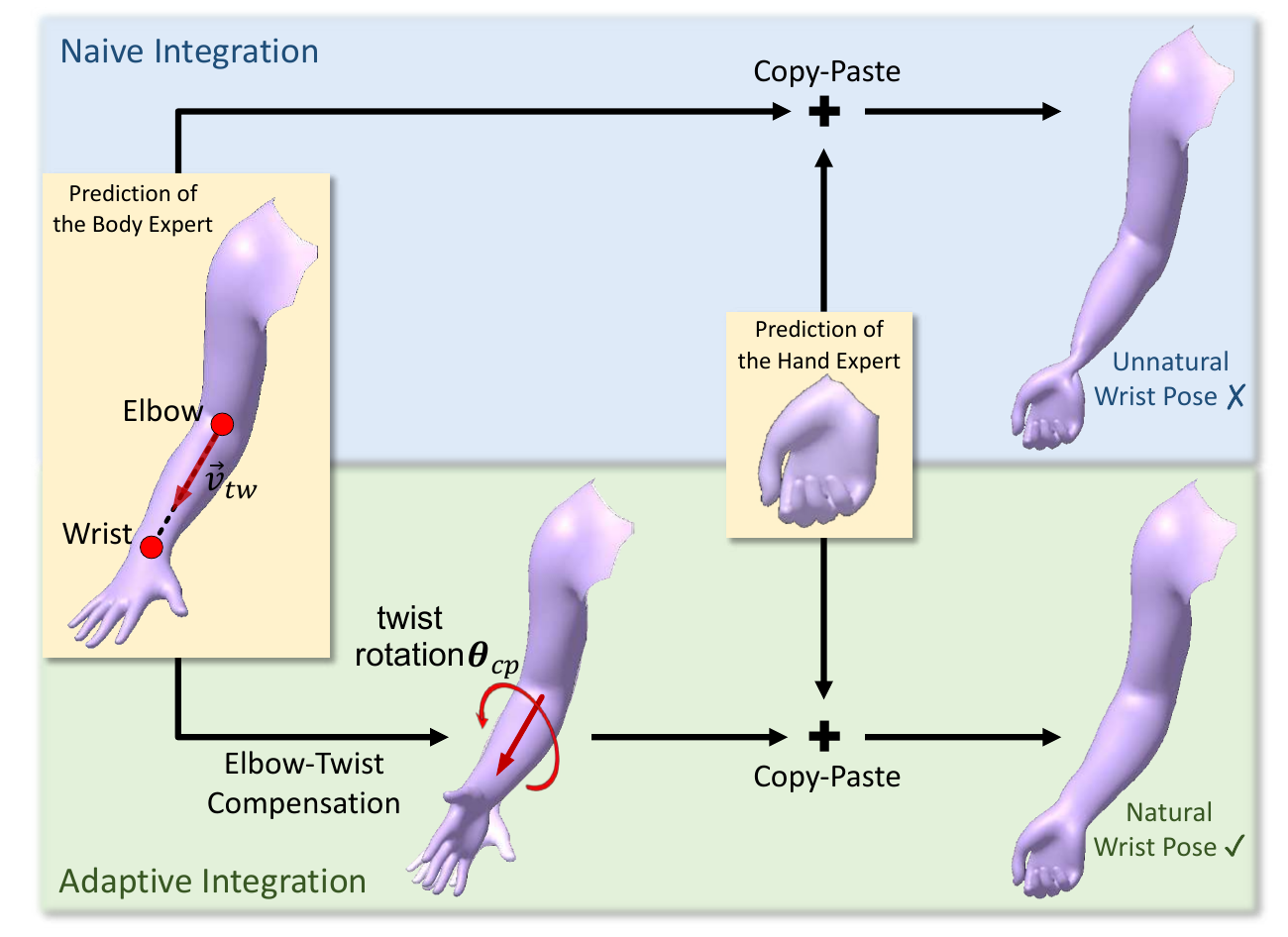}
		\vspace{-3mm}
		\caption{Illustration of the naive integration and the proposed adaptive integration with elbow-twist compensation.}
		\vspace{-5mm}
		\label{fig:elbow_twist}
	\end{center}
\end{figure}

\paragraph{Adaptive Integration with Elbow-Twist Compensation.}
As pointed out in previous work~\cite{choutas2020monocular}, the body expert hardly perceives the hand poses due to the small proportion of hand region in the body images.
It may lead to incompatible configurations of the arm and hand poses predicted individually by the body and hand experts, resulting in unnatural wrist poses of the full-body model, as illustrated in Fig.~\ref{fig:elbow_twist}.
Previous work~\cite{choutas2020monocular,rong2020frankmocap,feng2021collaborative,moon2022Hand4Whole} alleviates this issue by learning wrist poses from the body and hand features but typically degrades the accuracy of the wrist poses and alignment.
In our work, we propose an adaptive integration strategy to correct the elbow poses directly based on the solved wrist poses such that the elbow and wrist poses could be more compatible.
To maintain the mesh-image alignment, we only correct the twist rotation of the elbow joints as it is the rotation along the elbow-wrist bone and barely affects the position of the body and hand joints.
To this end, we first compute the twist angle of the wrist poses \wrt the elbow-to-wrist bone, then update the elbow and wrist poses by adding and subtracting the compensated twist rotation, respectively.

\textbf{Step 1: Computing the original twist angle.} The twist component around the elbow-to-wrist vector can be decomposed from the wrist poses. Without loss of generality, let the quaternion representation of the left or right wrist pose solved in Eq.~\eqref{eq:wrist} be $\bm{q}_{wrist}=(w_{wrist},\vec{\bm{v}}_{wrist})$.
By using Huyghe’s method~\cite{huyghe2011design,dobrowolski2015swing}, the quaternion $\bm{q}_{tw}$ of the twist rotation around the normalized elbow-to-wrist vector $\vec{\bm{v}}_{tw}$ can be calculated as:
\begin{equation}
\begin{aligned}
u_{proj} &= \frac{\vec{\bm{v}}_{wrist}\cdot \vec{\bm{v}}_{tw}}{||\vec{\bm{v}}_{wrist}||},\\
\bm{q}_{proj} &= (w_{wrist}, u_{proj}\vec{\bm{v}}_{tw}),\\
\bm{q}_{tw} &= \frac{\bm{q}_{proj}}{||\bm{q}_{proj}||},
\end{aligned}
\end{equation}
where $u_{proj}\vec{\bm{v}}_{tw}$ in $\bm{q}_{proj}$ is the projection vector of the normalized $\vec{\bm{v}}_{wrist}$ onto $\vec{\bm{v}}_{tw}$.
Let $w_{tw}$ be the first element of the twist quaternion $\bm{q}_{tw}$, then the twist rotation angle can be computed as $\alpha_{tw}=2cos^{-1}(w_{tw})\in[-\pi, \pi]$.

\textbf{Step 2: Updating elbow and wrist poses.} The angle $\alpha_{tw}$ reflects the intensity of the wrist rotation around the elbow-to-wrist bone, and an out-of-range twist angle typically leads to unnatural wrist poses.
To tackle this issue, an additional twist rotation is added to the elbow pose and serves as an compensation to the wrist pose.
Specifically, the elbow/wrist poses are updated as $\tilde{\bm{\theta}}_{elbow}/\tilde{\bm{\theta}}_{wrist}$ by adding/subtracting a twist rotation $\bm{\theta}_{cp}$ around the elbow-to-wrist vector $\vec{\bm{v}}_{tw}$ with a compensation angle of $\alpha_{cp}$, \ie, $\tilde{\bm{\theta}}_{elbow}=\bm{\theta}_{elbow}\bm{\theta}_{cp}$ and $\tilde{\bm{\theta}}_{wrist}=\bm{\theta}_{cp}^{-1}\bm{\theta}_{wrist}$.
In our solution, we empirically set a range $[\alpha_{tmin}, \alpha_{tmax}]$ to constraint $\alpha_{tw}$ and adopt the compensation angle $\alpha_{cp}$ as:
\begin{equation}
\alpha_{cp} = 
\begin{cases}
    \alpha_{tw}-\alpha_{tmax}, & \text{if } \alpha_{tw} > \alpha_{tmax},\\
    min(\alpha_{tw}-\alpha_{tmin},0), & \text{otherwise}.
\end{cases}
\end{equation}

As shown in our experiments, with the twist compensation from the elbow joint, the wrist pose becomes more natural while maintaining the mesh-image alignment of the body and hands.
In practice, the adaptive integration is not applied for those invisible hands since the global orientation predicted by the hand expert is not reliable when the hand is invisible.
In our implementation, the hand expert of PyMAF-X also predicts the confidence of the visibility status of hands.
When the hand is invisible, the full-body model simply adopts the wrist poses predicted by the body expert and the mean poses of hands.

%% file: tex/4_Experiments.tex
\section{Experiments}\label{sec:experiments}

\subsection{Implementation Details}

The part-specific PyMAF primarily adopts ResNet-50~\cite{he2016deep} as the backbone of the image encoder.
We also follow ExPose~\cite{choutas2020monocular} and PIXIE~\cite{feng2021collaborative} to adopt HRNet-W48\cite{sun2019deep} as the encoder backbone for the body model regression.
For each part-specific PyMAF, the image encoder takes a $224\times224$ image as input and produces spatial feature maps with resolutions of $\{14\times 14, 28\times 28, 56\times 56\}$.
When generating mesh-aligned features, the vertex number of body, hand, and face meshes is down-sampled to $431$, $195$, and $708$, respectively.
The mesh-aligned features extracted from feature maps of each point will be processed by MLPs so that their channel dimensions will be reduced to $5$.
Hence, the mesh-aligned feature vector for the body model has a length of $2155 = 431 \times 5$, which is similar to the length of the global features used in HMR~\cite{kanazawa2018end}.
The maximum number $T$ is set to $3$, which is equal to the iteration number used in HMR.
For the grid features used at $t=0$, they are uniformly sampled from $\bm{\phi}_s^0$ with a $21\times21$ grid pattern, \ie, the point number is $441 = 21\times21$ which is approximate to the vertices number $431$ after mesh downsampling.
The regressors $\mathcal{R}_t$ have the same architecture as the one in HMR, except that they have slightly different input dimensions.
The twist angle constraint $[\alpha_{tmin}, \alpha_{tmax}]$ is empirically set to $[-72\degree, 72\degree]$ in our implementation.
Following previous work~\cite{kolotouros2019learning,kocabas2021pare}, we adopt the continuous 6D representation~\cite{zhou2019continuity} for pose parameters in the regressor.
Following PARE~\cite{kocabas2021pare}, the body encoder is initialized with the model pretrained on 2D pose datasets~\cite{andriluka20142d,lin2014microsoft}.
During training, we use the Adam~\cite{kingma2014adam} optimizer and set the learning rate to $5e{-}5$ without decay.
The part-specific PyMAFs are first pre-trained individually and then assembled together for finetuning on full-body datasets.
Similar to PARE~\cite{kocabas2021pare}, we also observed a slight performance gain when removing the auxiliary supervision at the final stage of training, but we do not apply such a strategy in our experiments for more consistent ablation studies of our newly introduced components.
More details of the implementation can be found in our code and the Supplementary Material.

\textbf{Camera Setting.}
We follow previous work~\cite{kolotouros2019learning} to use a weak perspective camera with a pre-defined focal length of 5,000 by default for training and evaluation.
When running experiments on AGORA~\cite{patel2021agora}, we use a perspective camera with the focal lengths estimated by SPEC~\cite{kocabas2021spec} as there are stronger perspective distortions in this dataset.
Incorporating the camera setting of SPEC~\cite{kocabas2021spec} with our method for more accurate mesh recovery is left for future work.

\textbf{Runtime.} The PyTorch implementation of the body-only PyMAF takes about 22 ms to process one sample on the machine with an NVIDIA RTX 3090 GPU.
For full-body mesh recovery, PyMAF-X takes about 80 ms to process one sample, which is on par with existing regression-based approaches~\cite{choutas2020monocular,feng2021collaborative,moon2022Hand4Whole}.
In our current implementation, the part-specific backbone networks run in sequence to process the body, hand, and face images. Running them in parallel would further reduce the runtime.

\addtolength{\tabcolsep}{-5pt}
\begin{table}[t]
  \centering
  \footnotesize	
  \caption{Reconstruction errors on 3DPW~\cite{von2018recovering} and Human3.6M~\cite{ionescu2014human3}. Backbone architectures are highlighted in the brackets.}
  \vspace{-3mm}
    \begin{tabular}{cl|ccc|cc}
    \toprule
    \multicolumn{2}{c|}{\multirow{2}[4]{*}{Method}} & \multicolumn{3}{c|}{3DPW} & \multicolumn{2}{c}{Human3.6M} \\
\cmidrule{3-7}    \multicolumn{2}{c|}{} & PVE   & MPJPE & PA-MPJPE & MPJPE & PA-MPJPE \\
    \midrule
    \multirow{7}[2]{*}{\begin{sideways}Temporal\end{sideways}} & Kanazawa \etal~\cite{kanazawa2019learning} & 139.3 & 116.5 & 72.6  & -     & 56.9 \\
          & Doersch \etal~\cite{doersch2019sim2real} & -     & -     & 74.7  & -     & - \\
          & Arnab \etal~\cite{arnab2019exploiting} & -     & -     & 72.2  & 77.8  & 54.3 \\
          & DSD~\cite{sun2019human} & -     & -     & 69.5  & 59.1  & 42.4 \\
          & VIBE~\cite{kocabas2020vibe} & 113.4 & 93.5  & 56.5  & 65.9  & 41.5 \\
          & MEVA~\cite{luo20203d} & -     & 86.9  & 54.7  & -     & - \\
          & TCMR~\cite{choi2021beyond} & 111.5 & 95.0 & 55.8  & 62.3  & 41.1 \\
    \midrule
    \multirow{10}[2]{*}{\begin{sideways}Multiple Stage\end{sideways}} & Pavlakos \etal~\cite{pavlakos2018learning} & -     & -     & -     & -     & 75.9 \\
          & NBF~\cite{omran2018neural} & -     & -     & -     & -     & 59.9 \\
          & Zanfir \etal~\cite{zanfir2020weakly} & -     & 90.0  & 57.1  & -     & - \\
          & I2L-MeshNet~\cite{moon2020i2l} & -     & 93.2  & 58.6  & 55.7  & 41.1 \\
          & Pose2Mesh~\cite{choi2020pose2mesh} & -     & 88.9  & 58.3  & 64.9  & 46.3 \\
          & LearnedGD~\cite{song2020human} & -     & -     & 55.9  & -     & 56.4 \\
          & HUND~\cite{zanfir2021neural} & -     & 81.4  & 57.5  & 69.5  & 52.6 \\
          & HybrIK~\cite{li2021hybrik} & 94.5  & 80.0  & 48.8  & 54.4  & \textbf{34.5} \\
          & Pose2Pose~\cite{moon2022Hand4Whole} & - & 86.6  & 54.4  & -     & - \\
          & 3DCrowdNet~\cite{choi20213dcrowdnet} & 98.3 & 81.7  & 51.5  & -     & - \\
    \midrule
    \multirow{19}[4]{*}{\begin{sideways}Single Stage\end{sideways}} & HMR (Res50)~\cite{kanazawa2018end} & -     & 130.0 & 76.7  & 88.0  & 56.8 \\
          & GraphCMR (Res50)~\cite{kolotouros2019convolutional} & -     & -     & 70.2  & -     & 50.1 \\
          & SPIN (Res50)~\cite{kolotouros2019learning} & 116.4 & 96.9  & 59.2  & 62.5  & 41.1 \\
          & HMR-EFT (Res50)~\cite{joo2021exemplar} & -     & -     & 54.3  & -     & 46.0 \\
          & Graphormer (HR64)~\cite{lin2021mesh} & -     & -     & -     & \textbf{51.2}  & \textbf{34.5} \\
          & ROMP (Res50)~\cite{sun2021monocular} & 105.6 & 89.3  & 53.5  & -     & - \\
          & PARE (Res50)~\cite{kocabas2021pare} & 99.7  & 82.9  & 52.3  & -     & - \\
          & PARE (HR32)~\cite{kocabas2021pare} & 97.9  & 82.0  & 50.9  & -     & - \\
          \cdashline{2-7}[1.5pt/3pt]
          & Baseline (Res50) & 99.8  & 84.4  & 51.3  & 63.6  & 44.7 \\
          & PyMAF (Res50) & 94.4  & 79.7  & 49.0  & 58.1  & 40.2 \\
          & Baseline (HR48) & 93.3  & 79.5  & 48.0  & 58.8  & 39.5 \\
          & PyMAF (HR48) & \textbf{91.3} & \textbf{78.0} & \textbf{47.1} & 54.2  & 37.2 \\
\cmidrule{2-7}          & \textbf{* with training on 3DPW} &       &       &       &       &  \\
          & * HMR-EFT (Res50)~\cite{joo2021exemplar} & -     & -     & 51.6  & -     & - \\
          & * ROMP (Res50)~\cite{sun2021monocular} & 94.7  & 79.7  & 49.7  & -     & - \\
          & * Graphormer (HR64)~\cite{lin2021mesh} & 87.7  & 74.7  & 45.6  & -     & - \\
          & * PARE (HR32)~\cite{kocabas2021pare} & 88.6  & 74.5  & 46.5  & -     & - \\
          \cdashline{2-7}[1.5pt/3pt]
          & * PyMAF (Res50) & 88.7  & 76.8  & 46.8  & -     & - \\
          & * PyMAF (HR48) & \textbf{87.0} & \textbf{74.2} & \textbf{45.3} & -     & - \\
    \bottomrule
    \end{tabular}%
    \vspace{-5mm}
  \label{tab:3dpose}%
\end{table}%
\addtolength{\tabcolsep}{5pt}

\subsection{Datasets}
Following the practices of previous work~\cite{kanazawa2018end,kolotouros2019learning,kocabas2021pare}, the body expert is trained on a mixture of data from several datasets with 3D and 2D annotations, including Human3.6M~\cite{ionescu2014human3}, MPI-INF-3DHP~\cite{mehta2017monocular}, MPII~\cite{andriluka20142d}, LSP~\cite{johnson2010clustered}, LSP-Extended~\cite{johnson2011learning}, and COCO~\cite{lin2014microsoft}.
For the hand expert, we use images from FreiHAND~\cite{zimmermann2019freihand}, InterHand2.6M~\cite{moon2020interhand2} and COCO-Wholebody~\cite{jin2020whole} for training.
For the face expert, we use the images from VGGFace2~\cite{Cao2018_VGGFace2} for training.
Detailed descriptions of the datasets can be found in the Supplementary Material.

\paragraph{Pseudo Ground-truth.}
Following previous work~\cite{kocabas2021pare}, the SMPL/SMPL-X models fitted in EFT\cite{joo2021exemplar} and ExPose\cite{choutas2020monocular} are used as pseudo ground-truth annotations for the training of body and full-body model regressors.
For the training of the face expert, we use DECA~\cite{DECA_2020} and a face alignment algorithm FAN~\cite{bulat2017far} to generate pseudo ground-truth FLAME models and facial landmarks on the training set of VGGFace2~\cite{Cao2018_VGGFace2}.

\paragraph{Dense Correspondence.}
Note that we do not use the DensePose annotations in COCO for auxiliary supervision but render dense correspondence maps based on the pseudo ground-truth meshes using the method described in~\cite{zhang2020learning}.

\subsection{Evaluation Metrics}
\label{sec:metric}

We report the results of our approach in various evaluation metrics for quantitative comparisons with existing state-of-the-art methods, where all metrics are computed in the same way as previous work~\cite{kanazawa2018end,kolotouros2019learning,kocabas2021pare,choutas2020monocular,feng2021collaborative,moon2022Hand4Whole} in literature.

To quantitatively evaluate the performance of the 3D pose estimation, PVE, MPJPE, PA-PVE, and PA-MPJPE are adopted as the primary evaluation metrics.
They are all reported in millimeters (mm) by default.
Among these metrics, PVE denotes the mean Per-vertex Error, defined as the average point-to-point Euclidean distance between the predicted and ground truth mesh vertices, while MPJPE denotes the Mean Per Joint Position Error.
PA-PVE and PA-MPJPE denote the PVE and MPJPE after rigid alignment of the prediction with the ground truth using Procrustes Analysis.
Note that the metrics PA-PVE and PA-MPJPE are not aware of the global rotation and scale errors since they are calculated after rigid alignment.

\subsection{Comparison with the State of the Art}

\subsubsection{Evaluation on Body-only Mesh Recovery}

\paragraph{3D Human Pose and Shape Estimation.}
We first evaluate our approach on the 3D human pose and shape estimation task and make comparisons with previous state-of-the-art regression-based methods.
We present evaluation results for quantitative comparison on 3DPW~\cite{von2018recovering} and Human3.6M~\cite{ionescu2014human3} datasets in Table~\ref{tab:3dpose}.
Our PyMAF achieves competitive or superior results among previous approaches, including frame-based and temporal approaches.
Note that the approaches reported in Table~\ref{tab:3dpose} are not strictly comparable since they may use different training data, pseudo ground-truths, learning rate schedules, training epochs, \etc.
For a fair comparison, we report our baseline results in Table~\ref{tab:3dpose}, which is trained under the same setting as PyMAF.
The baseline approach has the same network architecture with HMR~\cite{kanazawa2018end} and also adopts the 6D rotation representation~\cite{zhou2019continuity} for pose parameters.
Under the setting of using ResNet-50 backbone and without training on 3DPW, PyMAF reduces the MPJPE over the baseline by 4.7 mm and 5.5 mm on 3DPW and Human3.6M datasets, respectively.

From Table~\ref{tab:3dpose}, we can see that PyMAF has more notable improvements on the metrics MPJPE and PVE.
We would argue that the metric PA-MPJPE can not reveal the mesh-image alignment performance since it is calculated as the MPJPE after rigid alignment.
As depicted in the Supplementary Material, a reconstruction result with a smaller PA-MPJPE value can have a larger MPJPE and worse alignment between the reprojected mesh and the input image.

\paragraph{2D Human Pose Estimation.}
We evaluate 2D human pose estimation performance on the COCO validation set to measure the mesh-image alignment quantitatively in real-world scenarios.
During the evaluation, we project keypoints from the estimated mesh on the image plane and compute the Average Precision (AP) based on the keypoint similarity with the ground truth 2D keypoints.
The results of keypoint localization APs are reported in Table~\ref{tab:coco}.
OpenPose~\cite{cao2019openpose}, a widely-used 2D human pose estimation algorithm, is also included for reference.
We can see that the COCO dataset is very challenging for approaches to human mesh recovery as they typically have much worse performances in terms of the 2D keypoint localization accuracy.
In Table~\ref{tab:coco}, we also include the results of the optimization-based SMPLify~\cite{bogo2016keep} by fitting the SMPL model to the ground-truth 2D keypoints with 1,500 optimization iterations.
As pointed out in previous work~\cite{kolotouros2019learning}, SMPLify may produce well-aligned but unnatural results.
Moreover, SMPLify is much more time-consuming than regression-based solutions.
Among approaches to recovering 3D human mesh, PyMAF outperforms previous regression-based methods by remarkable margins, making it the most competitive mesh recovery method in comparison with OpenPose~\cite{cao2019openpose}.
Under the backbone of ResNet-50, PyMAF brings significant improvements over our baseline by 8.5\% and 6.2\% on $\textrm{AP}$ and $\textrm{AP}_{50}$, respectively.
Qualitative comparisons can be found in the supplementary material.

\begin{table}[t]
 \footnotesize
  \centering
    \caption{Keypoint localization APs on the COCO~\cite{lin2014microsoft} validation set. Results of SMPLify~\cite{bogo2016keep} are evaluated based on the implementation in SPIN~\cite{kolotouros2019learning}. Results of HMR~\cite{kanazawa2018end}, GraphCMR~\cite{kolotouros2019convolutional}, and SPIN~\cite{kolotouros2019learning} are evaluated based on their publicly released code and models.}
    \vspace{-3mm}
    \begin{tabular}{l|ccccc}
    \toprule
    Method & AP    & $\textrm{AP}_{50}$ & $\textrm{AP}_{75}$ & $\textrm{AP}_{M}$ & $\textrm{AP}_{L}$ \\
    \midrule
    OpenPose~\cite{cao2019openpose} & \textbf{65.3} & \textbf{85.2} & \textbf{71.3} & \textbf{62.2} & \textbf{70.7} \\
    \midrule
    SMPLify~\cite{bogo2016keep} & 29.0  & 47.5  & 30.0  & 35.6  & 24.3 \\
    HMR~\cite{kanazawa2018end} & 18.9  & 47.5  & 11.7  & 21.5  & 17.0 \\
    CMR~\cite{kolotouros2019convolutional} & 9.3   & 26.9  & 4.2   & 11.3  & 8.1 \\
    SPIN~\cite{kolotouros2019learning} & 17.3  & 39.1  & 13.5  & 19.0  & 16.6 \\
    DaNet~\cite{zhang2020learning} & 33.8  & 68.6  & 29.9  & 36.0  & 32.3 \\
    \midrule
    Baseline (Res50) & 33.7  & 65.8  & 31.5  & 36.3  & 31.9 \\
    PyMAF (Res50) & 42.2  & 72.0  & 44.6  & 44.4  & 41.0 \\
    Baseline (HR48) & 44.9  & 74.8  & 48.4  & 48.1  & 42.8 \\
    PyMAF (HR48) & \textbf{47.7} & \textbf{76.7} & \textbf{52.5} & \textbf{50.5} & \textbf{46.1} \\
    \bottomrule
    \end{tabular}%
  \label{tab:coco}%
\end{table}%

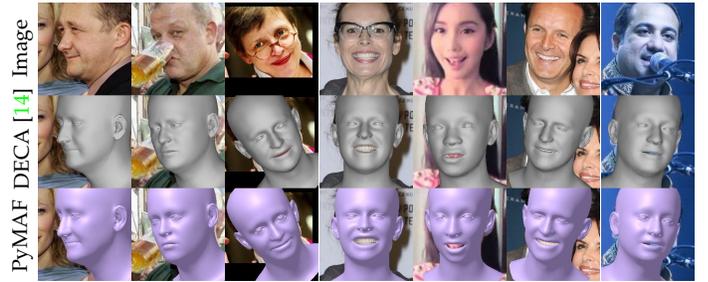
\begin{figure}[t]
	\raggedleft
    \begin{tikzpicture}[remember picture,overlay]
	\node[font=\fontsize{8pt}{8pt}\selectfont, rotate=90] at (-0.1,3.1) {Image};
    \node[font=\fontsize{8pt}{8pt}\selectfont, rotate=90] at (-0.1,1.9) {DECA~\cite{DECA_2020}};
 	\node[font=\fontsize{8pt}{8pt}\selectfont, rotate=90] at (-0.1,0.6) {PyMAF};
	\end{tikzpicture}
	\includegraphics[width=0.48\textwidth]{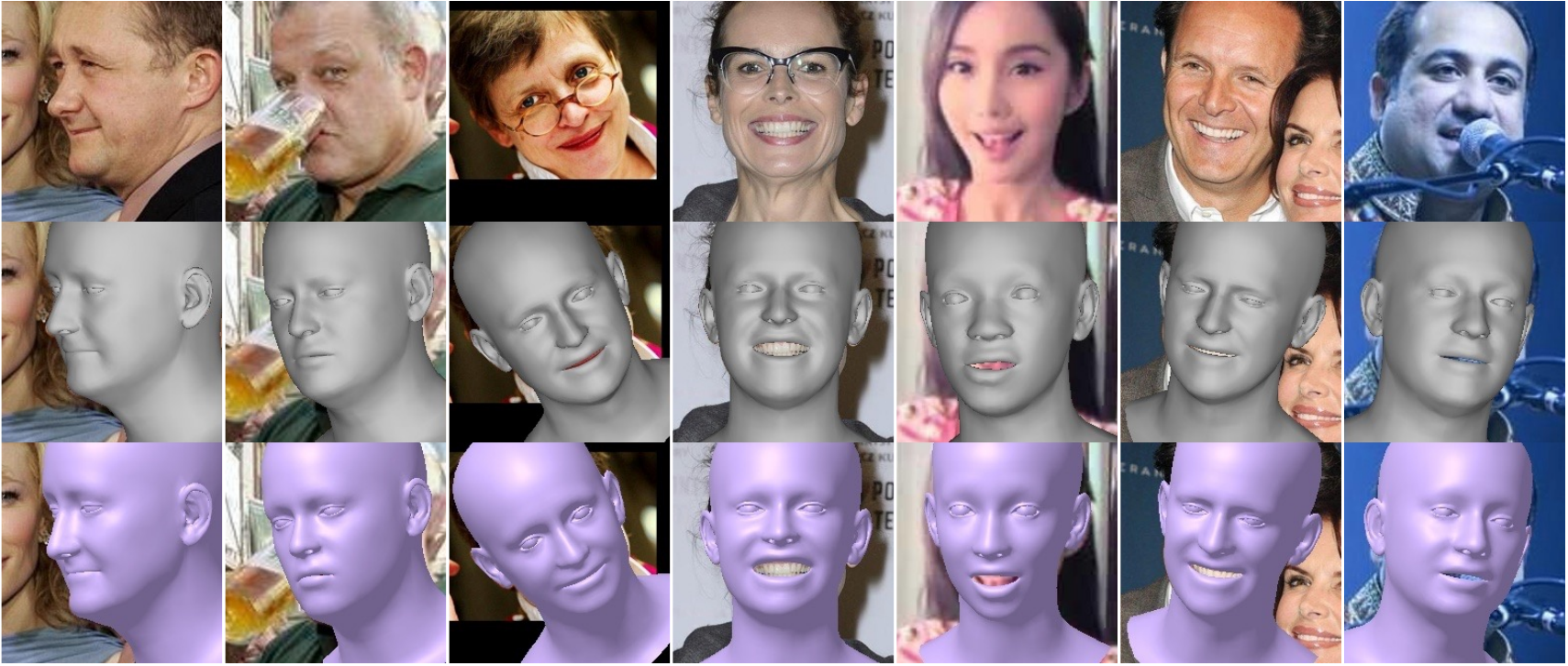}
	\vspace{-7mm}
	\caption{Qualitative comparison of face reconstruction results on in-the-wild images. The results of both DECA~\cite{DECA_2020} and PyMAF are the parametric FLAME~\cite{li2017learning} models.}
	\vspace{-3mm}
	\label{fig:demo_face}
\end{figure}

\begin{table}[t]
  \centering
  \caption{Hand reconstruction errors on the FreiHAND~\cite{zimmermann2019freihand} dataset. $^\dagger$ denotes the methods using extra training data more than FreiHAND.}
  \vspace{-2mm}
    \begin{tabular}{lccc}
    \toprule
    \multicolumn{1}{c}{Method} & PA-PVE $\downarrow$ & PA-MPJPE $\downarrow$ & \makecell{F-Scores $\uparrow$ @ \\ 5 mm / 10 mm} \\
    \midrule
    \textbf{* Hand-only methods} &       &       &  \\
    FreiHAND~\cite{zimmermann2019freihand} & 10.7 & -     & 0.529 / 0.935 \\
    Pose2Mesh~\cite{choi2020pose2mesh} & 7.8  & 7.7  & 0.674 / 0.969 \\
    I2L-MeshNet~\cite{moon2020i2l} & 7.6  & 7.4  & 0.681 / 0.973 \\
    METRO~\cite{lin2021end} & \textbf{6.7} & \textbf{6.8} & \textbf{0.717 / 0.981} \\
    \midrule
    \textbf{* Full-body methods} &       &       &  \\
    ExPose~\cite{choutas2020monocular} & 11.8 & 12.2 & 0.484 / 0.918 \\
    Zhou et al.~\cite{zhou2021monocular} & -     & 15.7  & - / - \\
    FrankMocap~\cite{rong2020frankmocap} & 11.6 & 9.2  & 0.553 / 0.951 \\
    PIXIE~\cite{feng2021collaborative} $^\dagger$ & 12.1 & 12.0 & 0.468 / 0.919 \\
    Hand4Whole~\cite{moon2022Hand4Whole} $^\dagger$ & 7.7  & \textbf{7.7} & 0.664 / 0.971 \\
    \hdashline[1.5pt/3pt]
    Baseline & 8.6   & 8.9   & 0.605 / 0.963 \\
    PyMAF & 8.1   & 8.4   & 0.638 / 0.969 \\
    PyMAF $^\dagger$ & \textbf{7.5} & \textbf{7.7} & \textbf{0.671 / 0.974} \\
    \bottomrule
    \end{tabular}%
    \vspace{-5mm}
  \label{tab:hand_eval}%
\end{table}%

{
\begin{table}[htbp]
  \centering
  \caption{Face reconstruction errors on Stirling3D~\cite{Feng2018evaluation} and NoW~\cite{Sanyal2019_ringnet} datasets.}
  \vspace{-2mm}
    \begin{tabular}{lccc}
    \toprule
    \multicolumn{1}{c}{\multirow{2}[4]{*}{Method}} & \multicolumn{3}{c}{PA-P2S (mm) $\downarrow$} \\
\cmidrule{2-4}          & Median & Mean & Std \\
    \midrule
    \multicolumn{4}{l}{\textbf{* Stirling3D LQ/HQ}} \\
    RingNet~\cite{Sanyal2019_ringnet} & 1.63/1.58 & 2.08/2.02 & 1.79/1.69 \\
    ExPose~\cite{choutas2020monocular} & 1.76/1.91 & 2.27/2.42 & 1.97/2.03 \\
    \hdashline[1.5pt/3pt]
    Baseline & 1.55/1.57 & 2.02/2.04 & 1.77/1.78 \\
    PyMAF & \textbf{1.51/1.48} & \textbf{1.97/1.92} & \textbf{1.72/1.67} \\
    \midrule
    \multicolumn{4}{l}{\textbf{* NoW}} \\
    RingNet~\cite{Sanyal2019_ringnet} & 1.21 & 1.54 & 1.31 \\
    DECA~\cite{DECA_2020} & \textbf{1.09} & \textbf{1.38} & \textbf{1.18} \\
    ExPose~\cite{choutas2020monocular} & 1.26 & 1.57 & 1.32 \\
    PIXIE~\cite{feng2021collaborative} & 1.18 & 1.49 & 1.25 \\
    \hdashline[1.5pt/3pt]
    Baseline & 1.17 & 1.47 & 1.23 \\
    PyMAF & 1.13  & 1.42 & 1.20 \\
    \bottomrule
    \end{tabular}%
    \vspace{-5mm}
  \label{tab:face_only_eval}%
\end{table}%
}

\begin{figure*}[htbp]
	\raggedleft
	\begin{tikzpicture}[remember picture,overlay]
	\node[font=\fontsize{8pt}{8pt}\selectfont, rotate=90] at (-0.1,9) {Image};
 	\node[font=\fontsize{8pt}{8pt}\selectfont, rotate=90] at (-0.1,7.2) {PyMAF-X};
 	\node[font=\fontsize{8pt}{8pt}\selectfont, rotate=90] at (-0.1,5) {Hand4Whole~\cite{moon2022Hand4Whole}};
	\node[font=\fontsize{8pt}{8pt}\selectfont, rotate=90] at (-0.1,3) {PIXIE~\cite{feng2021collaborative}};
	\node[font=\fontsize{8pt}{8pt}\selectfont, rotate=90] at (-0.1,1.0) {FrankMocap~\cite{rong2020frankmocap}};
	\end{tikzpicture}
	\includegraphics[width=0.98\textwidth]{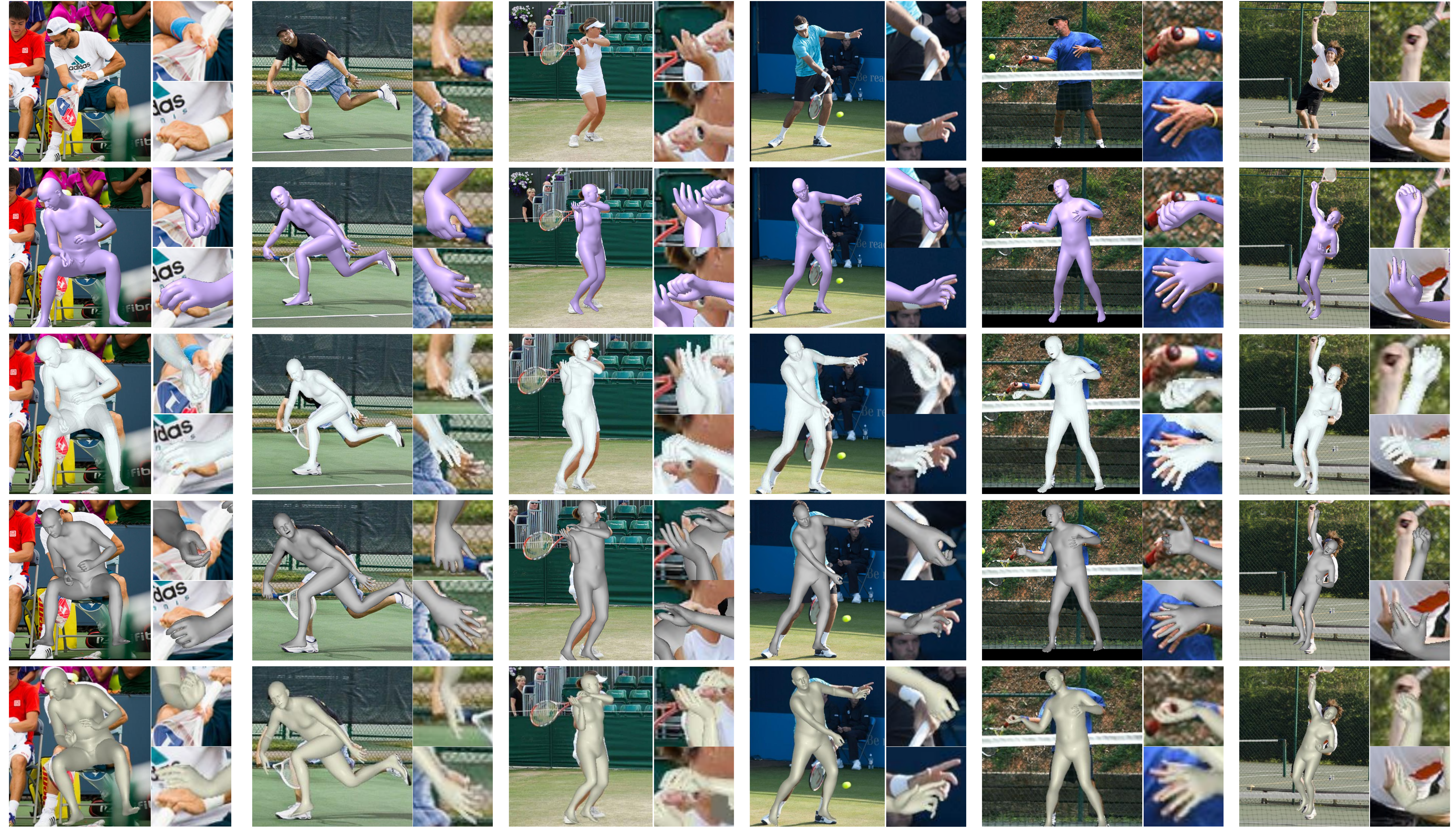}
	\vspace{-3mm}
	\caption{Qualitative comparison of full-body model reconstruction results on the COCO~\cite{lin2014microsoft} validation set.}
	\vspace{-5mm}
	\label{fig:demo_smplx}
\end{figure*}

\begin{table*}[t]
  \centering
  \caption{Reconstruction errors on the EHF~\cite{pavlakos2019expressive} test set. $^\dagger$ denotes the optimization-based approaches.}
  \vspace{-2mm}
    \begin{tabular}{l|ccc|cccc|cc}
    \toprule
    \multicolumn{1}{c|}{\multirow{2}[4]{*}{\textbf{Method}}} & \multicolumn{3}{c|}{\textbf{PVE}} & \multicolumn{4}{c|}{\textbf{PA-PVE}} & \multicolumn{2}{c}{\textbf{PA-MPJPE}} \\
\cmidrule{2-10}          & \textbf{Full-body} & \textbf{Hands} & \textbf{Face} & \textbf{Full-body} & \textbf{Body} & \textbf{Hands} & \textbf{Face} & \textbf{Body} & \textbf{Hands} \\
    \midrule
    MTC~\cite{xiang2019monocular} $^\dagger$ & -     & -     & -     & 67.2  & -     & -     & -     & 107.8 & 16.7 \\
    SMPLify-X~\cite{pavlakos2019expressive} $^\dagger$ & -     & -     & -     & 65.3  & 75.4  & 12.3  & 6.3   & 87.6  & 12.9 \\
    ExPose~\cite{choutas2020monocular} & 77.1  & 51.6  & 35.0  & 54.5  & 52.6  & 12.8  & 5.8   & 62.8  & 13.1 \\
    FrankMocap~\cite{rong2020frankmocap} & 107.6 & 42.8  & -     & 57.5  & 52.7  & 12.6  & -     & 62.3  & 12.9 \\
    Zhou \etal~\cite{zhou2021monocular} & 90.8  & 51.7  & 28.1  & 66.2  & 60.3  & 14.6  & 7.0   & 70.9  & 14.3 \\
    PIXIE~\cite{feng2021collaborative} & 89.2  & 42.8  & 32.7  & 55.0  & 53.0  & 11.1  & \textbf{4.6} & 61.5  & 11.6 \\
    Hand4Whole~\cite{moon2022Hand4Whole} & 76.8  & 39.8  & 26.1  & 50.3  & -     & 10.8  & 5.8   & 60.4  & 10.8 \\
    \midrule
    PyMAF-X (Res50) & 68.0  & 29.8  & 20.5  & \textbf{47.3} & 45.9  & \textbf{10.1} & 5.6   & 52.9  & \textbf{10.3} \\
    PyMAF-X (HR48) & \textbf{64.9} & \textbf{29.7} & \textbf{19.7} & 50.2  & \textbf{44.8} & 10.2  & 5.5   & \textbf{52.8} & \textbf{10.3} \\
    \bottomrule
    \end{tabular}%
    \vspace{-3mm}
  \label{tab:ehf_eval}%
\end{table*}%

\begin{table*}[t]
  \centering
  \caption[Reconstruction errors on the AGORA~\cite{patel2021agora} test set.]{Reconstruction errors on the AGORA test set. $^\dagger$ denotes the methods that are fine-tuned on the AGORA training set or similarly synthetic data~\cite{kocabas2021spec}. All results are taken from the official evaluation platform\footnotemark.}
  \vspace{-2mm}
    \begin{tabular}{l|c|c|cccc|cccc}
    \toprule
    \multicolumn{1}{c|}{\multirow{2}[4]{*}{\textbf{Method}}} & \multirow{2}[4]{*}{\textbf{\makecell{Body\\Model}}} & \textbf{Detection} & \multicolumn{4}{c|}{\boldmath{}\textbf{MVE $\downarrow$}\unboldmath{}} & \multicolumn{4}{c}{\boldmath{}\textbf{MPJPE $\downarrow$}\unboldmath{}} \\
\cmidrule{3-11}          &       & \boldmath{}\textbf{F1 Score $\uparrow$}\unboldmath{} & \textbf{Full-Body} & \textbf{Body} & \textbf{Face} & \textbf{L/R-Hand} & \textbf{Full-Body} & \textbf{Body} & \textbf{Face} & \textbf{L/R-Hand} \\
    \midrule
    SPIN~\cite{kolotouros2019learning} $^\dagger$ & SMPL  & 0.77  & -     & 148.9 & -     & -     & -     & 153.4 & -     & - \\
    PARE~\cite{kocabas2021pare} $^\dagger$ & SMPL  & 0.84  & -     & 140.9 & -     & -     & -     & 146.2 & -     & - \\
    SPEC~\cite{kocabas2021spec} $^\dagger$ & SMPL  & 0.84  & -     & 106.5 & -     & -     & -     & 112.3 & -     & - \\
    ROMP~\cite{sun2021monocular} $^\dagger$ & SMPL  & 0.91  & -     & 103.4 & -     & -     & -     & 108.1 & -     & - \\
    BEV~\cite{sun2021putting} $^\dagger$ & SMPL  & 0.93  & -     & 100.7 & -     & -     & -     & 105.3 & -     & - \\
    \midrule
    SMPLify-X~\cite{pavlakos2019expressive} & SMPL-X & 0.71  & 236.5 & 187.0 & 48.9  & 48.3/51.4 & 231.8 & 182.1 & 52.9  & 46.5/49.6 \\
    ExPose~\cite{choutas2020monocular} & SMPL-X & 0.82  & 217.3 & 151.5 & 51.1  & 74.9/71.3 & 215.9 & 150.4 & 55.2  & 72.5/68.8 \\
    FrankMocap~\cite{rong2020frankmocap} & SMPL-X & 0.80  & -     & 168.3 & -     & 54.7/55.7 & -     & 165.2 & -     & 52.3/53.1 \\
    PIXIE~\cite{feng2021collaborative} & SMPL-X & 0.82  & 191.8 & 142.2 & 50.2  & 49.5/49.0 & 189.3 & 140.3 & 54.5  & 46.4/46.0 \\
    Hand4Whole~\cite{moon2022Hand4Whole} $^\dagger$ & SMPL-X & \textbf{0.94} & 135.5 & 90.2  & 41.6  & 46.3/48.1 & 132.6 & 87.1  & 46.1  & 44.3/46.2 \\
    \midrule
    PyMAF-X (Res50) $^\dagger$ & SMPL-X & 0.89  & 134.4 & 90.4  & 38.7  & 45.9/47.3 & 132.8 & 89.0  & 42.2  & 43.9/45.4 \\
    PyMAF-X (HR48) $^\dagger$ & SMPL-X & 0.89  & \textbf{125.7} & \textbf{84.0} & \textbf{35.0} & \textbf{44.6/45.6} & \textbf{124.6} & \textbf{83.2} & \textbf{37.9} & \textbf{42.5/43.7} \\
    \bottomrule
    \end{tabular}%
    \vspace{-3mm}
  \label{tab:agora_eval}%
\end{table*}%

\subsubsection{Evaluation on Hand-only Reconstruction}

We compare the hand-only PyMAF with state-of-the-art approaches on the FreiHAND~\cite{zimmermann2019freihand} dataset.
As shown in Table~\ref{tab:hand_eval}, PyMAF outperforms the baseline and previous full-body methods and is comparable with recent hand-only methods~\cite{moon2020i2l,lin2021end}. It is also worth noting that full-body methods typically adopt the parametric representation of the hand mesh, which tends to be numerically inferior to the non-parametric representation used in recent hand-only methods~\cite{moon2020i2l,lin2021end}, as pointed out in previous works~\cite{moon2020i2l,tian2022recovering}.

{
\subsubsection{Evaluation on Face-only Reconstruction}

Following previous work~\cite{choutas2020monocular,feng2021collaborative}, we compare the face-only PyMAF with state-of-the-art face reconstruction approaches on the test set of Stirling3D~\cite{Feng2018evaluation} and NoW~\cite{Sanyal2019_ringnet} datasets.
Table~\ref{tab:face_only_eval} reports the performances of different methods in Point-to-Surface after Procrustes Alignment (PA-P2S).
It shows that the PyMAF outperforms the face expert of previous full-body methods ExPose~\cite{choutas2020monocular} and PIXIE~\cite{feng2021collaborative}, while achieving similar results compared with the strong face-only method DECA~\cite{DECA_2020}.
Qualitative comparisons of face reconstruction results are visualized in Fig.~\ref{fig:demo_face}.
For more consistent comparisons, we only show the intermediate parametric FLAME model predicted by DECA~\cite{DECA_2020} without using detail displacements.
We can see that PyMAF is able to capture expressive face shapes and has competitive results against DECA~\cite{DECA_2020}.
}

\subsubsection{Evaluation on Full-body Mesh Recovery}

Following previous work~\cite{choutas2020monocular,rong2020frankmocap,feng2021collaborative,moon2022Hand4Whole} on full-body mesh recovery, we evaluate the performance of PyMAF-X on two benchmark datasets, \ie, EHF~\cite{pavlakos2019expressive} and AGORA~\cite{patel2021agora}.

Table~\ref{tab:ehf_eval} reports the results of different methods for full-body mesh recovery, including the optimization-based MTC~\cite{xiang2019monocular} and SMPLify-X~\cite{pavlakos2019expressive}, and the regression-based ExPose~\cite{choutas2020monocular}, FrankMocap~\cite{rong2020frankmocap}, Zhou \etal~\cite{zhou2021monocular}, PIXIE~\cite{feng2021collaborative}, and Hand4Whole~\cite{moon2022Hand4Whole}. 
We can see that PyMAF-X achieves the best results among existing solutions on most metrics, especially on the evaluation of the body and full-body reconstruction.

Table~\ref{tab:agora_eval} compares the results of PyMAF-X and other full-body methods on the test set of AGORA~\cite{patel2021agora}, where all the evaluation results are taken from the official evaluation platform.
Recent state-of-the-art approaches to body-only mesh recovery are also included in Table~\ref{tab:agora_eval} for comprehensive comparisons.
Note that the evaluation on AGORA is also affected by the detection results as the predictions are first matched with the ground truth and then used to calculate the reconstruction error.
We use an off-the-shelf tool OpenPifPaf~\cite{kreiss2021openpifpaf} to detect persons and the corresponding hands and face regions, of which the person detection result is slightly worse than the recent solutions Hand4Whole~\cite{moon2022Hand4Whole} and BEV~\cite{sun2021putting}.
For matched predictions, PyMAF-X outperforms other methods, especially in the metrics for hand and full-body reconstruction on this challenging dataset.

Qualitative comparisons of different full-body methods on real-world images are shown in Fig.~\ref{fig:demo_smplx}, where we can see that PyMAF-X produces more accurate body, hand, and wrist poses than recent state-of-the-art approaches, including FrankMocap~\cite{rong2020frankmocap}, PIXIE~\cite{feng2021collaborative}, and Hand4Whole~\cite{moon2022Hand4Whole}. The video results of PyMAF-X and other full-body methods can be found on our project page and supplementary materials.

\subsection{Ablation Study}
In this part, we will perform ablation studies under various settings to validate the key components proposed in PyMAF and PyMAF-X.
The efficacy of the mesh-aligned features, pyramidal design, auxiliary dense supervision, and spatial alignment attention proposed in PyMAF will be validated on Human3.6M~\cite{ionescu2014human3}.
As the Human3.6M dataset includes large-scale amounts of images and the corresponding ground-truth 3D labels, ablation approaches of PyMAF are trained and evaluated on Human3.6M.
As for the proposed adaptive integration in PyMAF-X, ablation approaches are evaluated on EHF~\cite{pavlakos2019expressive}, where different approaches are trained under the same setting.

\paragraph{Efficacy of Mesh-aligned Features.}
In PyMAF, mesh-aligned features provide the current mesh-image alignment information in the feedback loop, which is essential for better mesh recovery.
To verify this, we alternatively replace the mesh-aligned features with the global features or the grid features uniformly sampled from spatial features as the input for parameter regressors.
Table~\ref{tab:MeshAligned} reports the performance of approaches using different types of features in the feedback loop.
The results under the non-pyramidal setting are also included in Table~\ref{tab:MeshAligned}, where the grid and mesh-aligned features are extracted from the feature maps with the highest resolution (\ie, $56 \times 56$), and the mesh-aligned features are extracted on the reprojected points of the mesh under the mean pose at $t=0$.
Note that all approaches in Table~\ref{tab:MeshAligned} do not use auxiliary supervision.

Unsurprisingly, using mesh-aligned features yields the best performance under both non-pyramidal and pyramidal designs.
The approach using the grid features sampled from spatial feature maps has better results than global features but is worse than the mesh-aligned counterpart.
The mesh-aligned solution achieves even more performance gain when using pyramidal feature maps since multi-scale mesh-alignment evidence is leveraged in the feedback loop.
Though the grid features contain primary spatial cues on uniformly distributed pixel positions, they can not reflect the alignment status of the current estimation.
This implies that mesh-aligned features are the most informative ones for the regressor to rectify the current mesh parameters.

\footnotetext{\url{https://agora-evaluation.is.tuebingen.mpg.de}}

\begin{table}[t]
  \centering
  \footnotesize
    \caption{Ablation study on using different types of feedback features for refinement. No auxiliary supervision is applied.}
    \vspace{-3mm}
    \begin{tabular}{l|c|cc}
    \toprule
    Feedback Feat. & Pyramid? & MPJPE & PA-MPJPE \\
    \midrule
    Global (Baseline) & \multirow{3}[2]{*}{No} & 84.1  & 55.6 \\
    Grid  &       & 80.5  & 54.7 \\
    Mesh-aligned &       & \textbf{79.6} & \textbf{53.4} \\
    \midrule
    Grid  & \multirow{2}[2]{*}{Yes} & 79.7  & 54.3 \\
    Mesh-aligned &       & \textbf{76.8}  & \textbf{50.9} \\
    \bottomrule
    \end{tabular}%
    \vspace{-3mm}
  \label{tab:MeshAligned}%
\end{table}%

\paragraph{Benefit from Auxiliary Supervision.}
The auxiliary pixel-wise supervision helps to enhance the reliability of the mesh-aligned evidence extracted from spatial features.
Using alternative pixel-wise supervision such as part segmentation rather than dense correspondences is also possible in our framework.
In our approach, these auxiliary predictions are solely needed for supervision during training since the point-wise features are extracted from feature maps.
For more in-depth analyses, we have also tried extracting point-wise features from the auxiliary predictions, \ie, the input type of regressors are intermediate representations such as part segmentation or dense correspondences.
Table~\ref{tab:AuxSupv} compares different auxiliary supervision settings and input types for regressors during training.
Using part segmentation is slightly worse than our dense correspondence solution.
Compared with the part segmentation, the dense correspondences preserve clean and rich information in foreground regions.
Moreover, using feature maps for point-wise feature extraction consistently performs better than auxiliary predictions.
This can be explained by the fact that using intermediate representations as input for regressors hampers the end-to-end learning of the whole network.
Under the auxiliary supervision strategy, the spatial feature maps are learned with the signal backpropagated from both auxiliary prediction and parameter correction tasks.
In this way, the background features can also contain information for mesh parameter correction since the deep features have larger receptive fields and are trained in an end-to-end manner.
As shown in Table~\ref{tab:AuxSupv}, when the mesh-aligned features are masked with the foreground region of part segmentation predictions, the performance degrades from 75.5 mm to 77.6 mm on MPJPE.

\begin{table}[t]
\footnotesize
  \centering
    \caption{Ablation study on using different auxiliary supervision settings and input types for regressors.}
    \vspace{-3mm}
    \begin{tabular}{c|c|cc}
    \toprule
    Aux. Supv. & Input Type & MPJPE & PA-MPJPE \\
    \midrule
    None  & Feature & 76.8  & 50.9 \\
    \midrule
    \multirow{3}[2]{*}{Part. Seg.} & Part. Seg. & 108.1 & 75.9 \\
          & Feature & 75.5  & 49.2 \\
          & Feature*Part. Seg. & 77.6  & 51.1 \\
    \midrule
    \multirow{2}[2]{*}{Dense Corr.} & Dense Corr. & 77.8  & 54.7 \\
          & Feature & \textbf{75.1} & \textbf{48.9} \\
    \bottomrule
    \end{tabular}%
    \vspace{-3mm}
  \label{tab:AuxSupv}%
\end{table}%

\paragraph{Efficacy of Spatial Alignment Attention.}
In our approach, Spatial Alignment Attention (SAA) is designed to enable the awareness of the whole image context in the regressor.
To validate its efficacy, we replace the spatial alignment attention with fully-connected layers to fuse the grid and mesh-aligned features.
As reported in Table~\ref{tab:abla_saa}, simply fusing the grid features (the second row) only brings marginal improvements in comparison with the approach using the spatial alignment attention (the third row).
The performances of PyMAF with or without spatial alignment attention across each refinement iteration are reported in Table~\ref{tab:iter}, where the PyMAF with spatial alignment attention improves the reconstruction results more quickly.

\begin{table}[t]
  \centering
  \caption{Ablation study on the spatial alignment attention.}
  \vspace{-3mm}
    \begin{tabular}{l|ccc}
    \toprule
    Feedback Feat. & PVE   & MPJPE & PA-MPJPE \\
    \midrule
    Mesh-aligned & 89.1  & 76.8  & 50.9 \\
     + Grid & 88.9  & 76.6  & 51.0 \\
     + SAA & \textbf{85.1} & \textbf{73.6} & \textbf{50.1} \\
    \bottomrule
    \end{tabular}%
    \vspace{-3mm}
  \label{tab:abla_saa}%
\end{table}%

\begin{table}[t]
  \centering
  \caption{Reconstruction errors across different iterations in the feedback loop. SAA denotes the spatial alignment attention.}
  \vspace{-3mm}
    \begin{tabular}{c|l|cccc}
    \toprule
    Method & \multicolumn{1}{c|}{Metric} & $M_0$ & $M_1$ & $M_2$ & $M_3$ \\
    \midrule
    \multirow{3}[2]{*}{PyMAF w /o SAA} & PVE   & 312.1 & 97.0  & 90.5  & 88.7 \\
          & MPJPE & 274.0 & 80.3  & 76.6  & 75.1 \\
          & PA-MPJPE & 131.7 & 52.1  & 49.9  & 48.9 \\
    \midrule
    \multirow{3}[2]{*}{PyMAF w. SAA} & PVE   & 312.1 & 91.2  & 83.6  & \textbf{81.8} \\
          & MPJPE & 274.0 & 78.2  & 73.2  & \textbf{72.1} \\
          & PA-MPJPE & 131.7 & 51.6  & 49.8  & \textbf{48.7} \\
    \bottomrule
    \end{tabular}%
  \label{tab:iter}%
\end{table}%

\begin{figure}[t]
	\raggedleft
	\begin{tikzpicture}[remember picture,overlay]
	\node[font=\fontsize{8pt}{8pt}\selectfont, rotate=90] at (-0.1,5.9) {Image};
 	\node[font=\fontsize{8pt}{8pt}\selectfont, rotate=90] at (-0.1,4.3) {Learned};
 	\node[font=\fontsize{8pt}{8pt}\selectfont, rotate=90] at (-0.1,2.6) {Copy-Paste};
 	\node[font=\fontsize{8pt}{8pt}\selectfont, rotate=90] at (-0.1,0.8) {Adaptive};
	\end{tikzpicture}
	\includegraphics[width=0.46\textwidth]{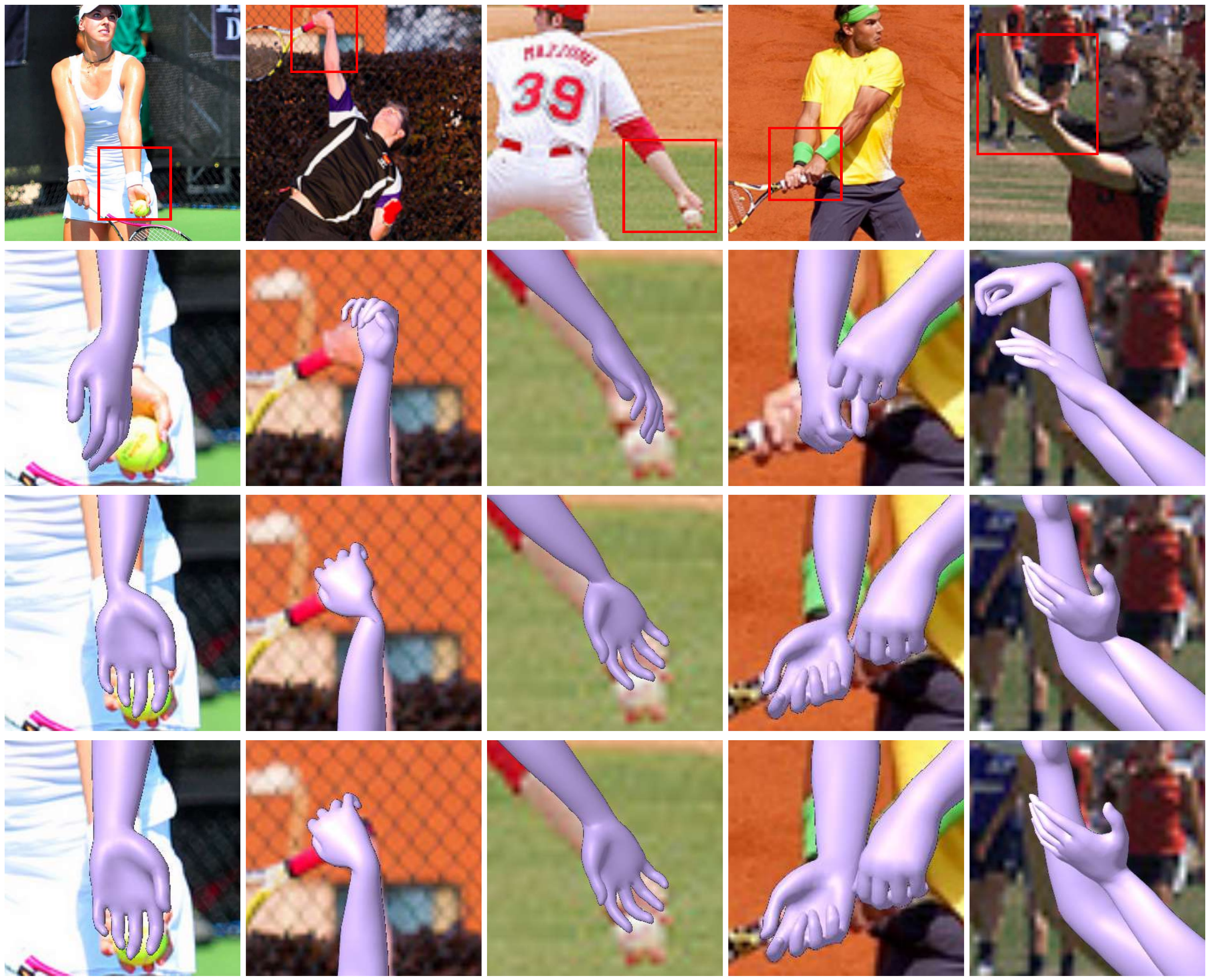}
	\vspace{-4mm}
	\caption{Visual comparison of different integration strategies under the cases of challenging wrist poses.}
	\vspace{-5mm}
	\label{fig:demo_twist}
\end{figure}
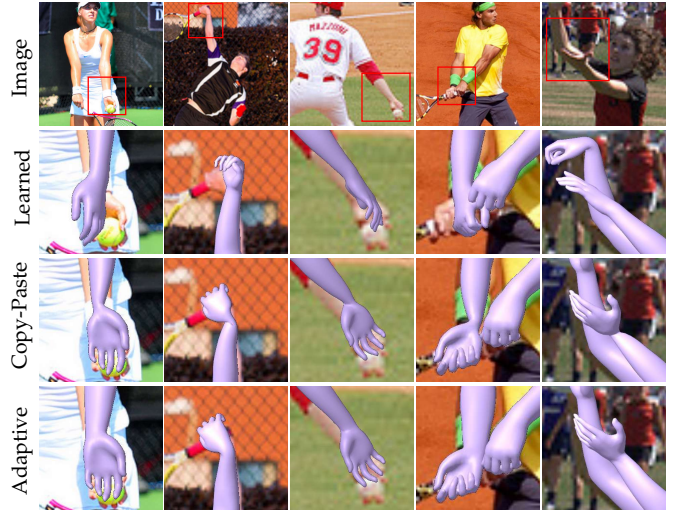

\paragraph{Efficacy of Adaptive Integration.}
In PyMAF-X, an elbow-twist compensation is used to adaptively correct the elbow poses in the integration of body and hand estimations.
Such an adaptive integration strategy could produce physically-plausible wrist poses while preserving the mesh-image alignment.
We investigate different integration strategies and compare our solution with two alternatives: i)
a learned integration strategy similar to PIXIE~\cite{feng2021collaborative}, which predicts the wrist poses based on the fused features of body and hand features; ii) the naive copy-paste integration strategy~\cite{rong2020frankmocap}, which calculates the wrist poses based on the estimated body and hand poses.
Table~\ref{tab:abla_integration} reports the performances of the three different integration strategies on the EHF dataset.
Here, we use the MPJPE of body and hand joints to measure the mesh-image alignment and the PA-PVE of wrist vertices to measure the physical plausibility of the wrist joint.
As shown in the first row, the learned integration strategy can also produce natural wrist poses but degrade the alignment of hand parts in the full-body model.
Compared with the learned and copy-paste strategies, the proposed adaptive integration produces both well-aligned and natural poses of the body, hand, and wrist parts.
Fig.~\ref{fig:demo_twist} provides a visual comparison of different integration strategies under challenging cases in real-world scenarios.
We can see that our adaptive integration maintains the alignment and effectively improves the plausibility of the wrist poses by leveraging the twist compensation from elbow joints.

\begin{table}[t]
  \centering
  \caption{Ablation study on the usage of the learned, copy-paste, and the proposed adaptive integration strategies.}
  \vspace{-3mm}
    \begin{tabular}{l|cc|c}
    \toprule
    \multicolumn{1}{c|}{\multirow{2}[2]{*}{\textbf{Integration}}} & \multicolumn{2}{c|}{\textbf{\makecell{MPJPE\\(Alignment)}}} & \textbf{\makecell{PA-PVE\\(Plausibility)}} \\
          & \textbf{Body} & \textbf{Hands} & \textbf{Wrist} \\
    \midrule
    Learned & \textbf{64.9} & 42.3  & \textbf{5.8} \\
    Copy-Paste & \textbf{64.9} & \textbf{31.2} & 6.7 \\
    Adaptive (Ours) & \textbf{64.9} & \textbf{31.2} & 5.9 \\
    \bottomrule
    \end{tabular}%
    \vspace{-3mm}
  \label{tab:abla_integration}%
\end{table}%

%% file: tex/5_Conclusion.tex
\section{Conclusion}\label{sec:conclusion}

In this paper, we first present Pyramidal Mesh Alignment Feedback (PyMAF) for regression-based human mesh recovery and further extend it as PyMAF-X for full-body mesh recovery.
PyMAF is primarily motivated by the observation of the reprojection misalignment between the parametric mesh results and the input images.
At the core of PyMAF, the parameter regressor leverages spatial information from a feature pyramid to correct the parameter deviation explicitly in a feedback loop based on the alignment status of the currently estimated meshes.
To achieve this, given a coarse-aligned mesh estimation, the mesh-aligned features are first extracted from the spatial feature maps and then fed back into the regressor for parameter rectification.
Moreover, an auxiliary dense supervision is employed to enhance the learning of mesh-aligned features while spatial alignment attention is introduced to enable the awareness of the global contexts in our deep regressor.
When extending PyMAF for full-body model recovery, an adaptive integration with the elbow-twist compensation strategy is proposed in PyMAF-X to produce natural wrist poses while maintaining the alignment performances of part-specific PyMAF.
The efficacy of PyMAF and PyMAF-X is validated on indoor and in-the-wild datasets, where our approaches effectively improve the mesh-image alignment over the baseline and previous regression-based solutions.

\paragraph{Limitations and Future Work.}
In our experiments, we found that PyMAF-X fails to reconstruct interacting hands due to the separated regression of two hands.
Meanwhile, when handling images with strong perspective distortions or with only upper-torso observations, common issues such as bent legs and erroneous limb poses remain unsolved in this work.
We will leave these issues for future work to incorporate the merits of recent datasets~\cite{muller2021self,huang2022capturing} and solutions such as SPEC~\cite{kocabas2021spec}, PIXIE~\cite{feng2021collaborative}, Hand4Whole~\cite{moon2022Hand4Whole}, and IntagHand~\cite{li2022interacting} into our framework.

Moreover, similar to existing methods~\cite{choutas2020monocular,rong2020frankmocap,feng2021collaborative,moon2022Hand4Whole}, the full-body alignment performance of PyMAF-X heavily relies on the pose and shape estimation of the body expert.
Due to the lack of full-body mesh annotations, the estimated body shapes are typically inaccurate in challenging cases, resulting in erroneous bone lengths of arms and coarse alignment of hands.
Combining PyMAF-X with SPIN~\cite{kolotouros2019learning}, EFT~\cite{joo2021exemplar}, or NeuralAnnot~\cite{moon2020neuralannot} for the generation of more precise pseudo 3D ground-truth full-body mesh annotations on in-the-wild data would be interesting future work.
Besides, the elbow-twist rotations are adjusted empirically in PyMAF-X based on the twist components of wrist poses.
Learning the compensation angle $\alpha_{cp}$ via networks is also possible when large-scale full-body mesh annotations are available.

%% file: tex/6_Appendix.tex
\section{More Implementation Details}

\paragraph{Attention Module.}
We follow Mesh Graphormer~\cite{lin2021mesh} to implement the attention module for the grid and mesh-aligned feature fusion.
We simply use a single attention block for each iteration because we found that a single block brought enough performance grains while using more attention blocks merely increased memory consumption.

\begin{figure}[t]
	\centering
	\includegraphics[width=0.5\textwidth]{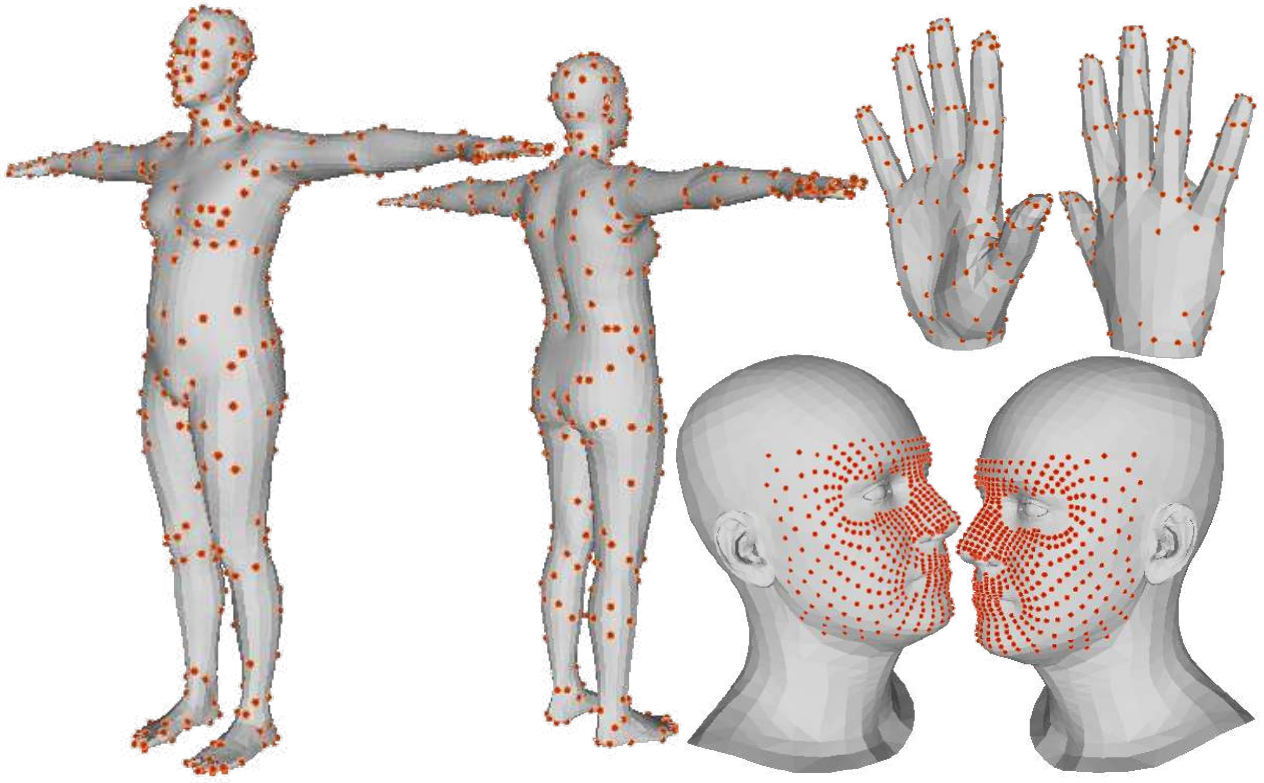}
    \vspace{-5mm}
	\caption{Visualization of the selected vertices on the body, hand, and face models for mesh-aligned feature extraction.}
	\label{fig:sampling_mesh}
\end{figure}

\paragraph{Mesh Sampling.}
For body and hand meshes, we simply use the down-sampling matrix provided in GraphCMR~\cite{kolotouros2019convolutional} and Mesh Graphormer~\cite{lin2021mesh} to reduce the vertex number for mesh-aligned feature extraction.
For face meshes, we manually select the vertices in the front face region considering that the expression information concentrates on the front face.
For body-only PyMAF, the dimension of the mesh-aligned features is reduced to match the feature dimension in HMR~\cite{kanazawa2018end} for more fair comparisons.
For hand- and face-only PyMAF, we do not strictly abide by this rule but simply replace SMPL with MANO/FLAME during the implementation.
Fig.~\ref{fig:sampling_mesh} visualizes the selected vertices on the body, hand, and face meshes.

\begin{figure}[t]
	\centering
	\includegraphics[width=0.5\textwidth]{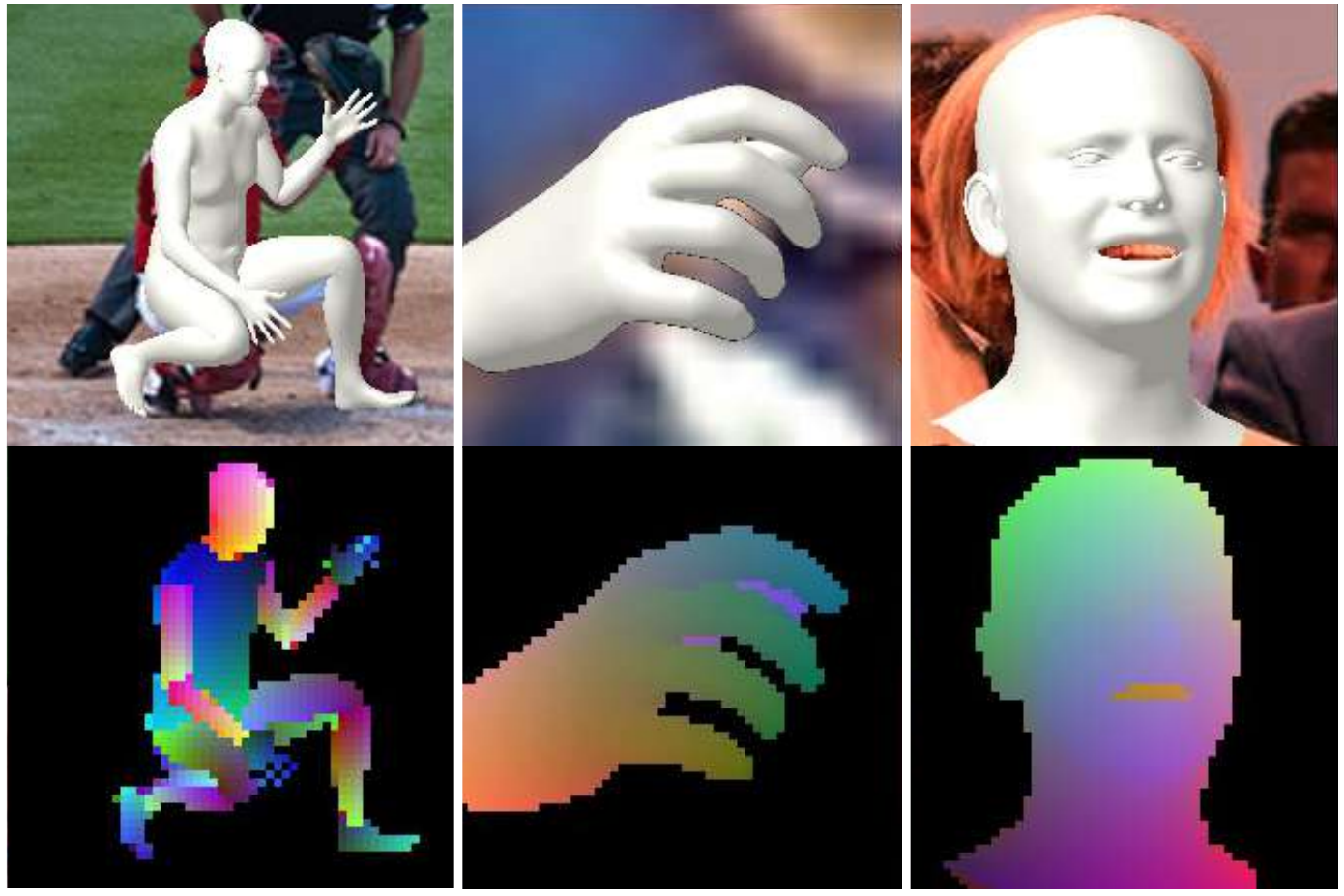}
    \vspace{-5mm}
	\caption{Visualization of the rendered dense correspondences in resolutions of $56 \times 56$ for the body, hand, and face meshes. From left to right: the IUV~\cite{guler2018densepose} map for a body model, the PNCC~\cite{zhu2017face} maps for hand and face models.}
    \vspace{-5mm}
	\label{fig:vis_dense_map}
\end{figure}

\paragraph{Auxiliary Dense Supervision.}
In the auxiliary dense prediction task, the dense correspondence and part segmentation (for ablation experiments) used for supervision are rendered based on the pseudo ground truth meshes.
Examples of the rendered dense correspondence are visualized in Fig.~\ref{fig:vis_dense_map}.
We choose to use the rendered dense correspondence mainly based on the following reasons:

i) The dense correspondence can be regarded as a more fine-grained part segmentation. In our method, we use the IUV representation as dense correspondence for the body expert, while using the Projected Normalized Coordinate Code (PNCC)~\cite{zhu2017face} for hand and face experts.
Dense correspondences are more general since PNCC does not need to split the mesh into manually defined parts.

ii) The body part definition may vary from different datasets while the rendered one is consistent. Moreover, the rendering process can be done efficiently in batch using the tools like PyTorch3D~\cite{ravi2020accelerating}. The costs of rendering part segmentation and dense correspondences are almost the same for the PyTorch3D renderer.

iii) There is only a limited number of datasets providing the annotated part segmentation and dense correspondence, while the pseudo ground truth meshes are typically available for common datasets thanks to previous work such as SPIN~\cite{kolotouros2019learning} and EFT~\cite{joo2021exemplar}.

\paragraph{Hand Visibility.}
We simply use three FC layers to predict hand visibility based on the hand-only mesh-aligned features. The pseudo visibility confidence used for supervision is calculated based on the proportion of the visible hand keypoints annotated in the whole-body COCO~\cite{jin2020whole} dataset.
For example, assume that there are 15 visible keypoints and the total number of hand keypoints is 21, then the pseudo visibility confidence of this hand is about 0.714 (15/21).

\section{About Metrics}
\begin{figure}[t]
	\centering
	\begin{subfigure}[b]{0.23\textwidth}
		\includegraphics[width=1\textwidth]{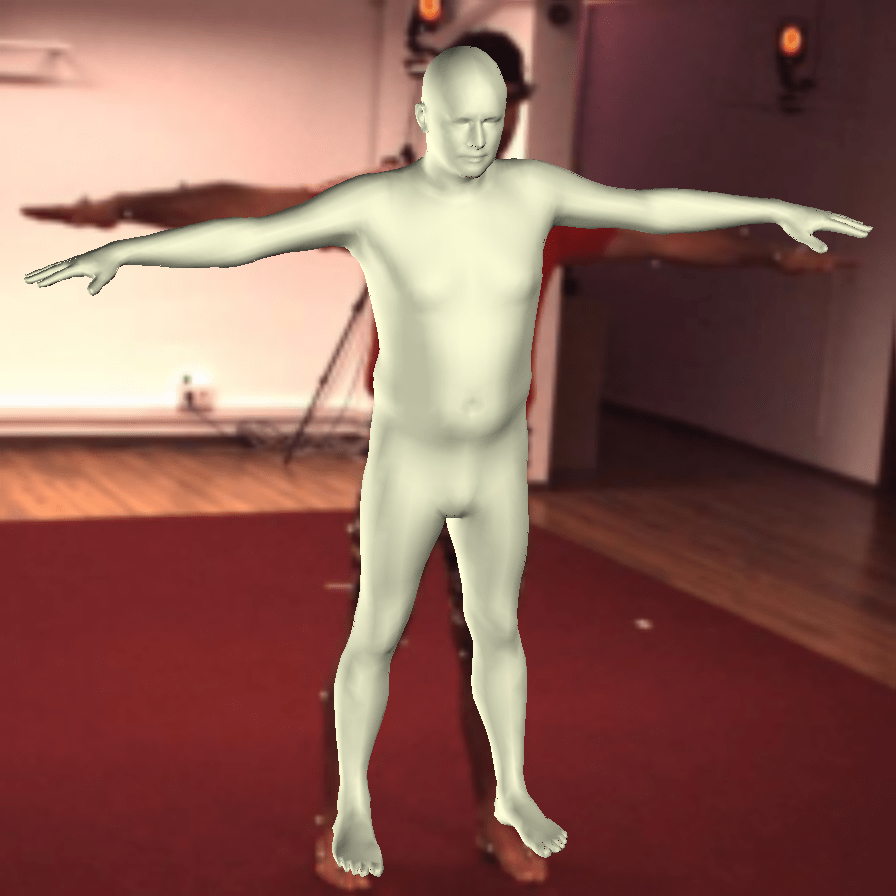}
		\caption{\scriptsize PA-MPJPE: 26.9, MPJPE: 74.3}
		\label{fig:bad}
    \end{subfigure}
	\begin{subfigure}[b]{0.23\textwidth}
		\includegraphics[width=1\textwidth]{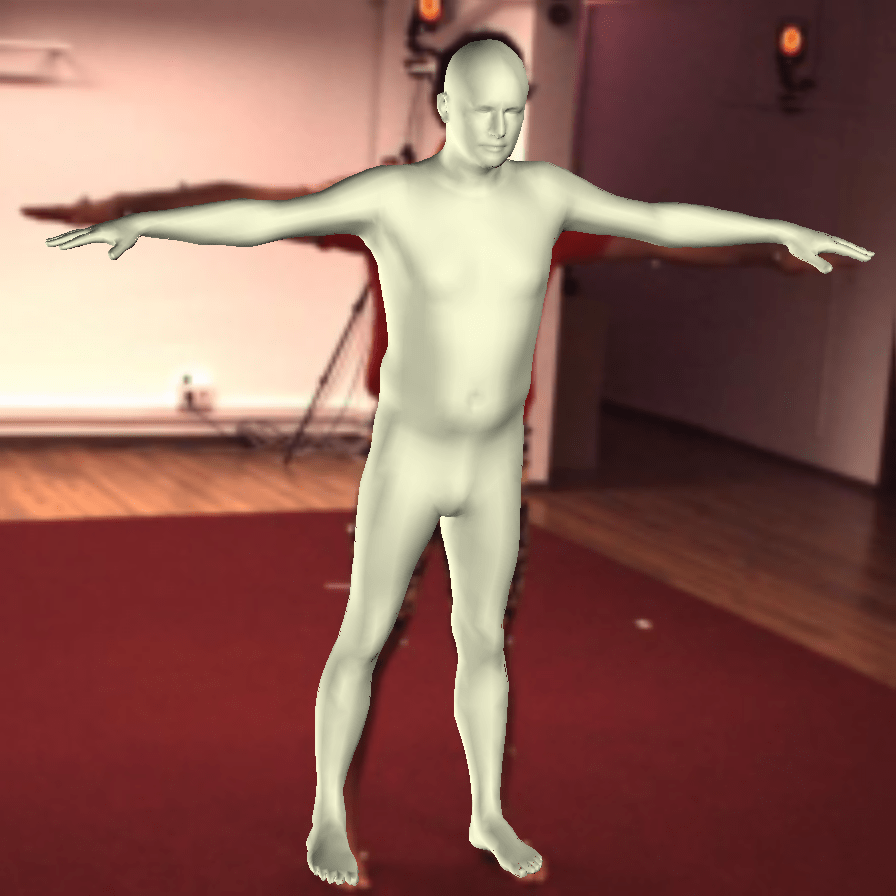}
		\caption{\scriptsize PA-MPJPE: 27.7, MPJPE: 43.4}
		\label{fig:good}
    \end{subfigure}
    \vspace{-2mm}
	\caption{Examples of two reconstruction results. (a) A reconstruction result with a lower PA-MPJPE value but worse mesh-image alignment. (b) A reconstruction result with a higher PA-MPJPE value but better mesh-image alignment.}
	\label{fig:metrics}
\end{figure}

Though the PA-PVE and PA-MPJPE are widely adopted in the 3D pose estimation task, these two metrics can not fully reveal the mesh-image alignment performance since they are calculated after rigid alignment.
As depicted in Fig.~\ref{fig:metrics}, a reconstruction result with a lower PA-MPJPE value can have a higher MPJPE value and worse alignment between the reprojected mesh and image.

\section{About Datasets}
\label{sec:datasets}
Following the practices of previous work~\cite{kanazawa2018end,kolotouros2019learning,kocabas2021pare}, we train our body model regression network on several datasets with 3D or 2D annotations, including Human3.6M~\cite{ionescu2014human3}, MPI-INF-3DHP~\cite{mehta2017monocular}, LSP~\cite{johnson2010clustered}, MPII~\cite{andriluka20142d}, COCO~\cite{lin2014microsoft}.
For hand-only and full-body model regression, FreiHAND~\cite{zimmermann2019freihand}, InterHand2.6M~\cite{moon2020interhand2}, FFHQ~\cite{karras2019style}, and COCO-WholeBody~\cite{jin2020whole} are also used for training.
Here, we provide more descriptions of the datasets to supplement the main manuscript.

\textbf{Human3.6M}~\cite{ionescu2014human3} is commonly used as the benchmark dataset for 3D human pose estimation, consisting of 3.6 million video frames captured in the controlled environment.
The ground truth SMPL parameters in Human3.6M are generated by applying MoSH~\cite{loper2014mosh} to the sparse 3D MoCap marker data, as done in Kanazawa \etal~\cite{kanazawa2018end}.
The original videos are down-sampled from 50fps to 10fps, resulting in 312,188 frames for training.
Following the common protocols~\cite{pavlakos2017coarse,pavlakos2018learning,kanazawa2018end}, our experiments use five subjects (S1, S5, S6, S7, S8) for training and two subjects (S9, S11) for evaluation.
The original videos are also down-sampled from 50 fps to 10 fps to remove redundant frames, resulting in 312,188 frames for training and 26,859 frames for evaluation.

\textbf{3DPW}~\cite{von2018recovering} is captured in challenging outdoor scenes with IMU-equipped actors under various activities.
This dataset provides accurate shape and pose ground truth annotations.
Following the protocol of previous work~\cite{kanazawa2019learning,kolotouros2019learning}, we do not use its data for training by default unless specified in the table.

\textbf{MPI-INF-3DHP}~\cite{mehta2017monocular} is a 3D human pose dataset covering more actor subjects and poses than Human3.6M.
The images of this dataset were collected under both indoor and outdoor scenes, and the 3D annotations were captured by a multi-camera marker-less MoCap system.
Hence, there is some noise in the 3D ground truth annotations.
The training set includes 8 subjects and there are 96,507 frames down-sampled from videos used for training.

\textbf{LSP}~\cite{johnson2010clustered} and \textbf{LSP-Extended}~\cite{johnson2011learning} are 2D human pose benchmark datasets, containing person images with challenging poses. There are 14 visible 2D keypoint locations annotated for each image and 10,428 samples used for training.

\textbf{MPII}~\cite{andriluka20142d} is a standard benchmark for 2D human pose estimation.
There are 25,000 images collected from YouTube videos covering a wide range of activities.
We discard those images without complete keypoint annotations, producing 14,667 samples for training.

\textbf{COCO}~\cite{lin2014microsoft} and \textbf{COCO-WholeBody}~\cite{jin2020whole} contain a large scale of person images labeled with 17 body keypoints, 42 hand keypoints, and 68 face keypoints.
We use COCO to train body-only PyMAF and leverage the hand keypoints in COCO-WholeBody during the training of hand- and face-only PyMAF and PyMAF-X.
Since this dataset does not contain ground-truth meshes, we conduct a quantitative evaluation on the 2D keypoint localization task using its validation set, which consists of 50,197 samples.

\textbf{EHF}~\cite{pavlakos2019expressive} contains 100 testing images of one subject captured in lab environments.
For each image, the corresponding 3D scans and ground-truth SMPL-X~\cite{pavlakos2019expressive} meshes are provided.
EHF is used for testing only and is commonly adopted as a full-body evaluation benchmark dataset in literature~\cite{choutas2020monocular,feng2021collaborative,moon2022Hand4Whole}.

\textbf{AGORA}~\cite{patel2021agora} is a synthetic dataset with accurate SMPL-X models fitted to 3D scans.
Since the ground-truth labels of its test set are not publicly available, the evaluation is performed on the official platform\footnote{\url{https://agora-evaluation.is.tuebingen.mpg.de}}.
For evaluation on AGORA, we use the training set of AGORA to finetune our model.

\textbf{FreiHAND}~\cite{zimmermann2019freihand} contains 130,240 samples for training and 3960 samples images for evaluation. For each sample in the training set, the MANO~\cite{romero2017embodied} parameters recovered from multi-view images are provided. We use this dataset for the training and evaluation of the hand expert.

\textbf{InterHand2.6M}~\cite{moon2020interhand2} is a large-scale real-captured hand dataset, providing accurate MANO parameters of interacting hands. We crop single-hand images from this dataset for the training of the hand expert.


{

\textbf{VGGFace2}~\cite{Cao2018_VGGFace2} is a large-scale face dataset. The images of this dataset are downloaded from Google and have large variations in pose, age, and ethnicity. It contains about 3 million images from training.
We run the method of FAN~\cite{bulat2017far} and DECA~\cite{DECA_2020} on its training set to generate the pseudo ground truth facial landmarks and FLAME~\cite{li2017learning} models for the training of the face expert.

\textbf{Stirling3D}~\cite{Feng2018evaluation} provides facial images with the ground-truth 3D scans. The test set contains 2,000 facial images in neutral expressions, including 1,344 low-quality (LQ) images and 656 high-quality (HQ) images. We follow previous work~\cite{choutas2020monocular,DECA_2020} to use it for evaluation only.

\textbf{NoW}~\cite{Sanyal2019_ringnet} contains the facial images captured with an iPhone X, and a separate 3D scan for each subject.
Its test set contains 1,702 images for evaluation.
Since the ground-truth scans of the test set are not publicly available, the evaluation is performed by following the instructions on the official website\footnote{\url{https://now.is.tue.mpg.de/index.html}}.
We follow previous work~\cite{choutas2020monocular,DECA_2020} to use this dataset for evaluation only.
}

\section{More Qualitative Results}
\label{sec:further_qualitative}

\begin{figure}[t]
	\centering
	\foreach \idx in {2,3,4,5} {
		\begin{subfigure}[h]{0.45\textwidth}
    		\includegraphics[width=1\textwidth]{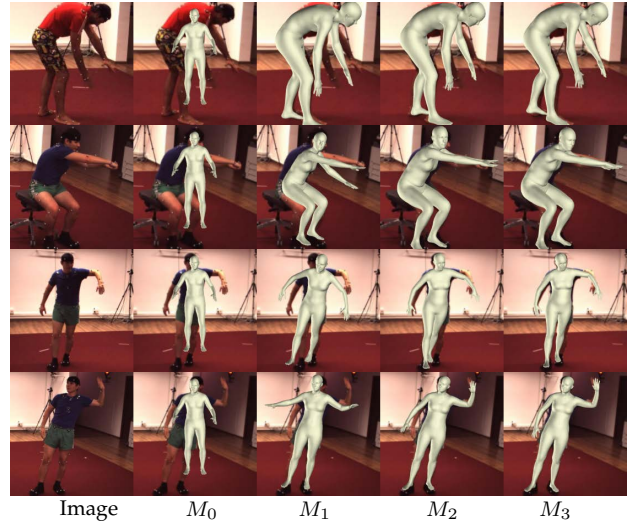}
		\end{subfigure}
	}
	\\
	\begin{tikzpicture}[remember picture,overlay]
	\node[font=\fontsize{8pt}{8pt}\selectfont] at (-3,-0.2) {Image};
	\node[font=\fontsize{8pt}{8pt}\selectfont] at (-1.5,-0.2) {$M_0$};
 	\node[font=\fontsize{8pt}{8pt}\selectfont] at (-0,-0.2) {$M_1$};
	\node[font=\fontsize{8pt}{8pt}\selectfont] at (1.7,-0.2) {$M_2$};
	\node[font=\fontsize{8pt}{8pt}\selectfont] at (3.2,-0.2) {$M_3$};
	\end{tikzpicture}
	\vspace{1mm}
	\caption{Visualization of reconstruction results across different iterations in the feedback loop.}
	\label{fig:loop}
\end{figure}

\begin{figure*}[t]
	\centering
	\foreach \idx in {1, 2, 3, 4, 5, 6, 7, 8, 9, 10, 11, 12, 13, 14, 15, 16} {
		\begin{subfigure}[h]{0.49\textwidth}
    		\includegraphics[width=1\textwidth]{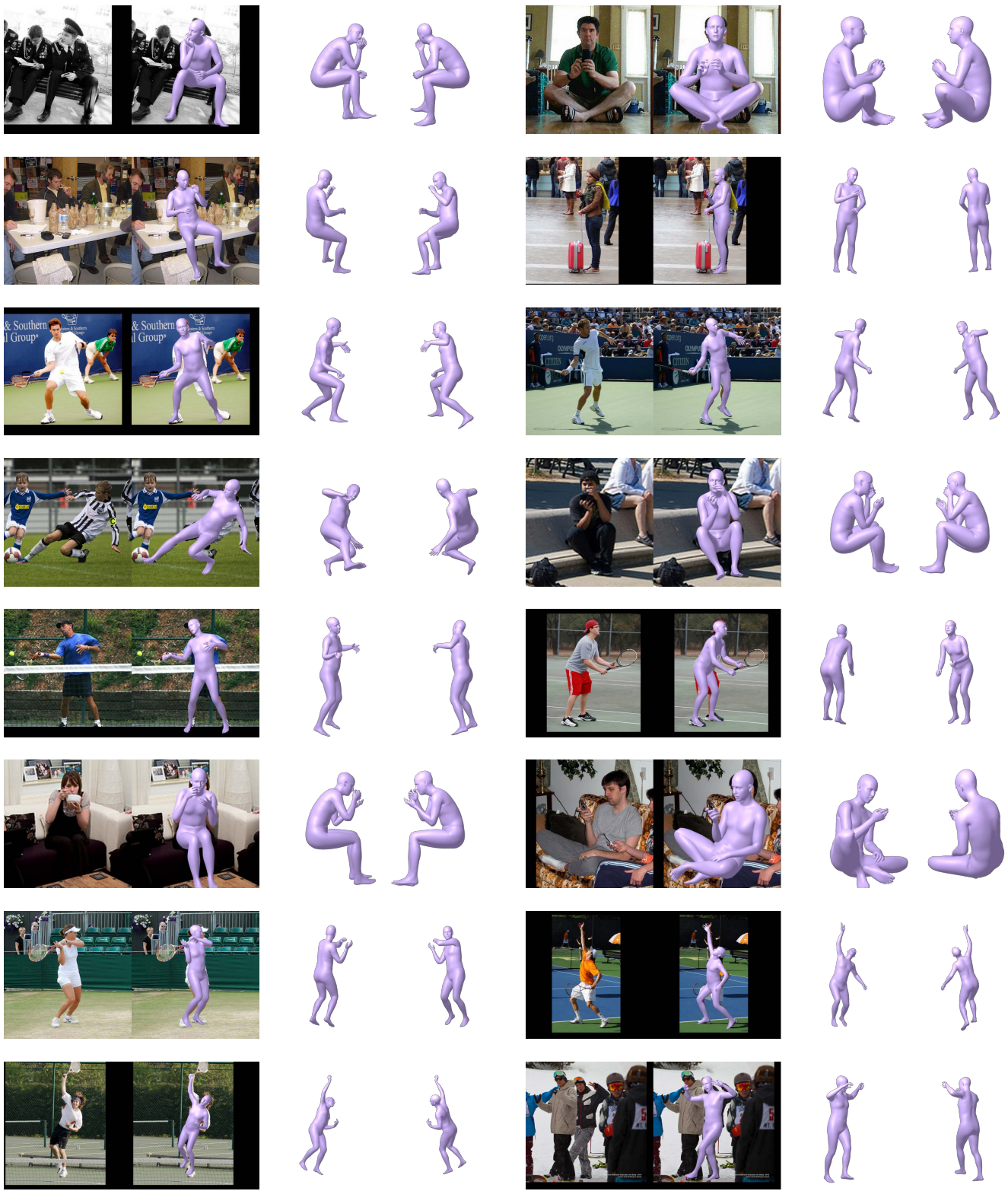}
      ~
		\end{subfigure}
	}
    \\
	\caption{PyMAF-X results visualized from different viewpoints. For each example, from left to right: the input image, the overlay result, and the results with rotations around the vertical axis. Best viewed zoomed-in on a color screen.}
	\vspace{-2mm}
	\label{fig:smplx_novel_view}
\end{figure*}

We provide more qualitative results of our method in this section.
In Fig~\ref{fig:loop}, we visualize the estimated meshes after each iteration, where it can be seen that PyMAF can correct the drift of body parts progressively and result in better-aligned human models.
In Fig.~\ref{fig:smpl_cocoDemo}, the body mesh recovery results of different methods on COCO are depicted for qualitative comparisons, where PyMAF convincingly performs better than competitors and our baseline by producing better-aligned and natural results.
In Fig.~\ref{fig:well-aligned}, we provide more full-body model reconstruction results on the COCO validation set, where PyMAF-X can produce well-aligned full-body model under challenging cases.
In Fig.\ref{fig:smplx_novel_view}, we further visualize the reconstructed full-body models from different viewpoints.

\begin{figure*}[t]
	\centering
	\foreach \idx in {1, 2, 3, 4, 5, 6, 7, 8, 9} {
		\begin{subfigure}[h]{0.32\textwidth}
    		\includegraphics[width=1\textwidth]{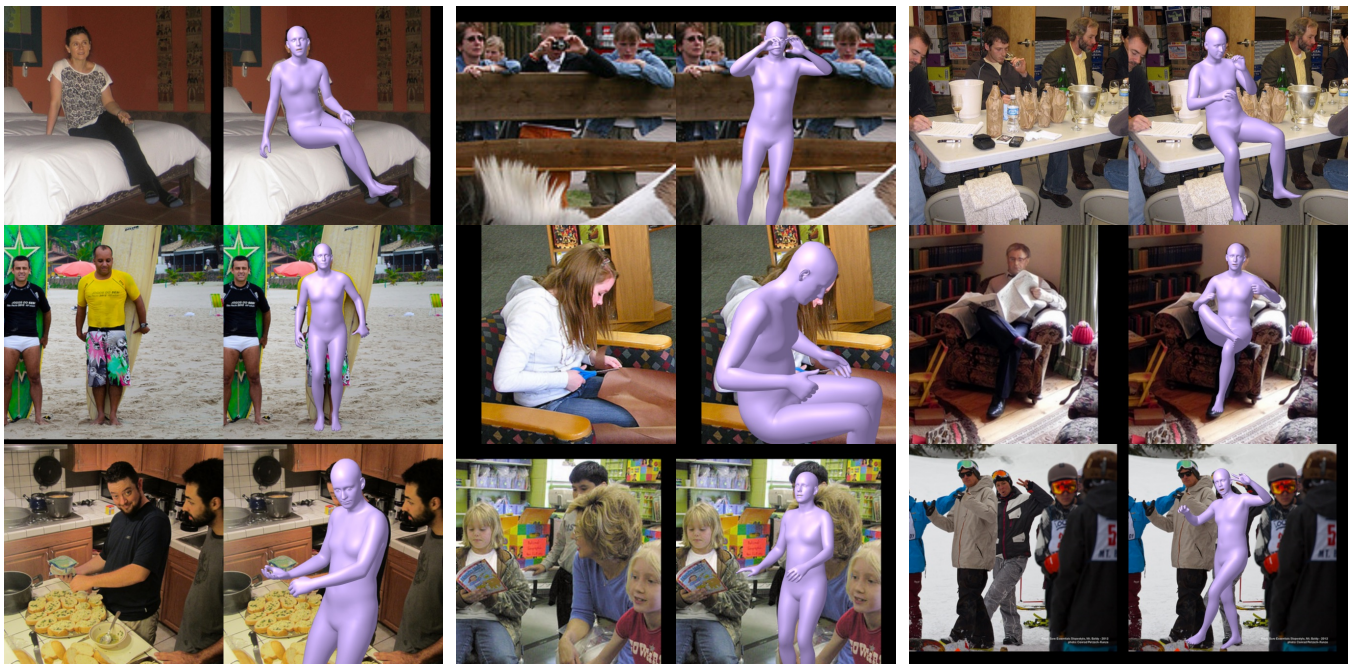}
		\end{subfigure}
	}
	\\
	\vspace{1mm}
	\caption{Example results of PyMAF-X when the body or hands are occluded. Samples come from the COCO validation set. Best viewed zoomed-in on a color screen.}
	\vspace{-5mm}
	\label{fig:occlusion}
\end{figure*}

\textbf{Cases under Occlusions.} As pointed out in the main paper, the adaptive integration is not applicable when the hand part is invisible. To handle this, the visibility status of hands is also predicted by the hand expert in PyMAF-X.
In cases of invisible hands, the full-body model adopts the default hand poses and the wrist poses estimated by the body expert.
Fig.~\ref{fig:occlusion} shows the example results of PyMAF-X when the body or hands are occluded.
We can see that PyMAF-X produces reasonable full-body meshes under these cases.

\begin{figure*}[t]
	\centering
    \begin{subfigure}[b]{0.49\textwidth}
		\includegraphics[width=1\textwidth]{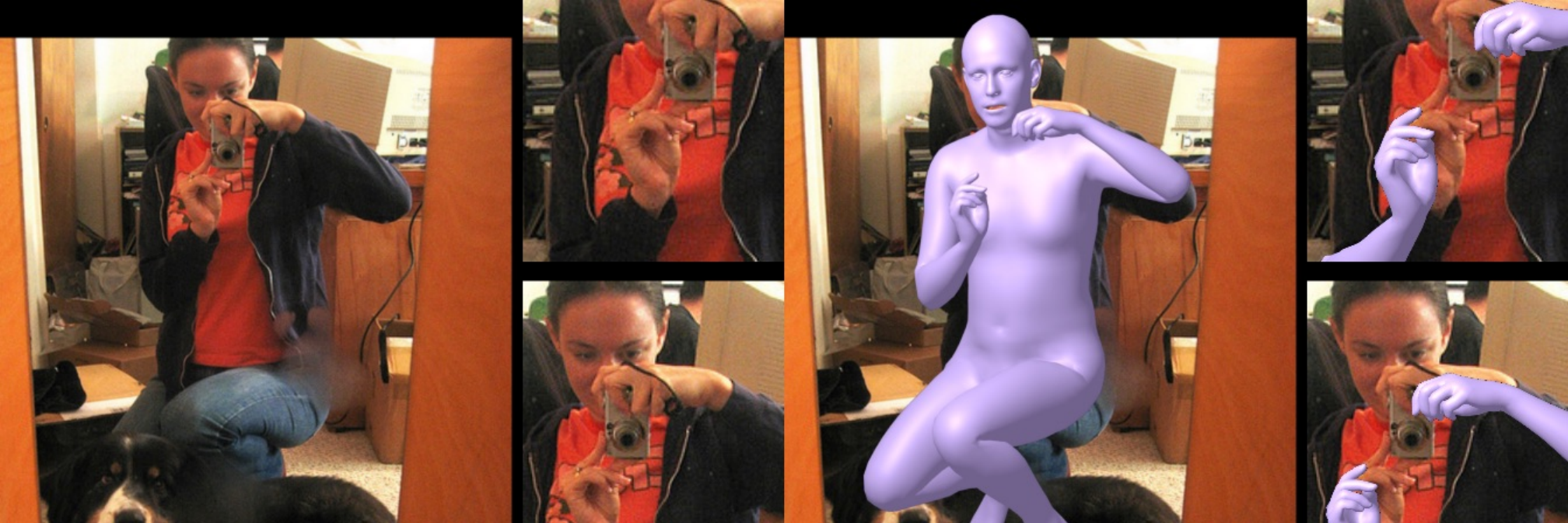}
		\caption{Inaccurate bone length (body shape)}
    \end{subfigure}
    \begin{subfigure}[b]{0.49\textwidth}
		\includegraphics[width=1\textwidth]{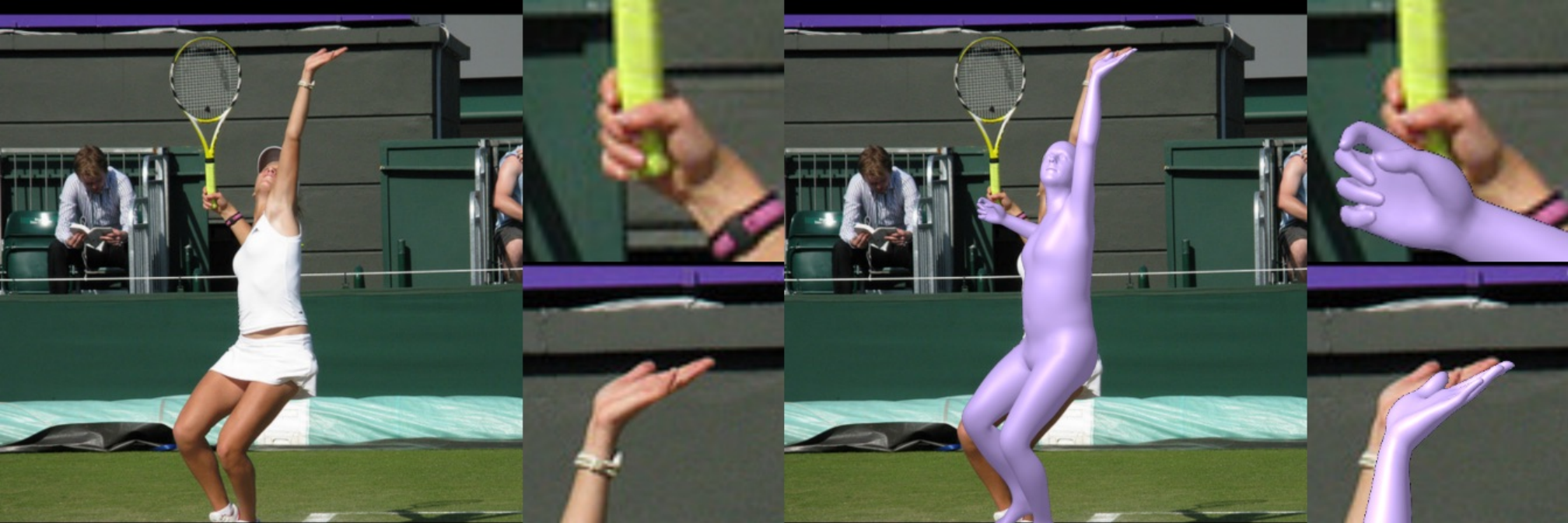}
		\caption{Inaccurate body pose}
    \end{subfigure}
    \\
    \vspace{2mm}
    \begin{subfigure}[b]{0.49\textwidth}
		\includegraphics[width=1\textwidth]{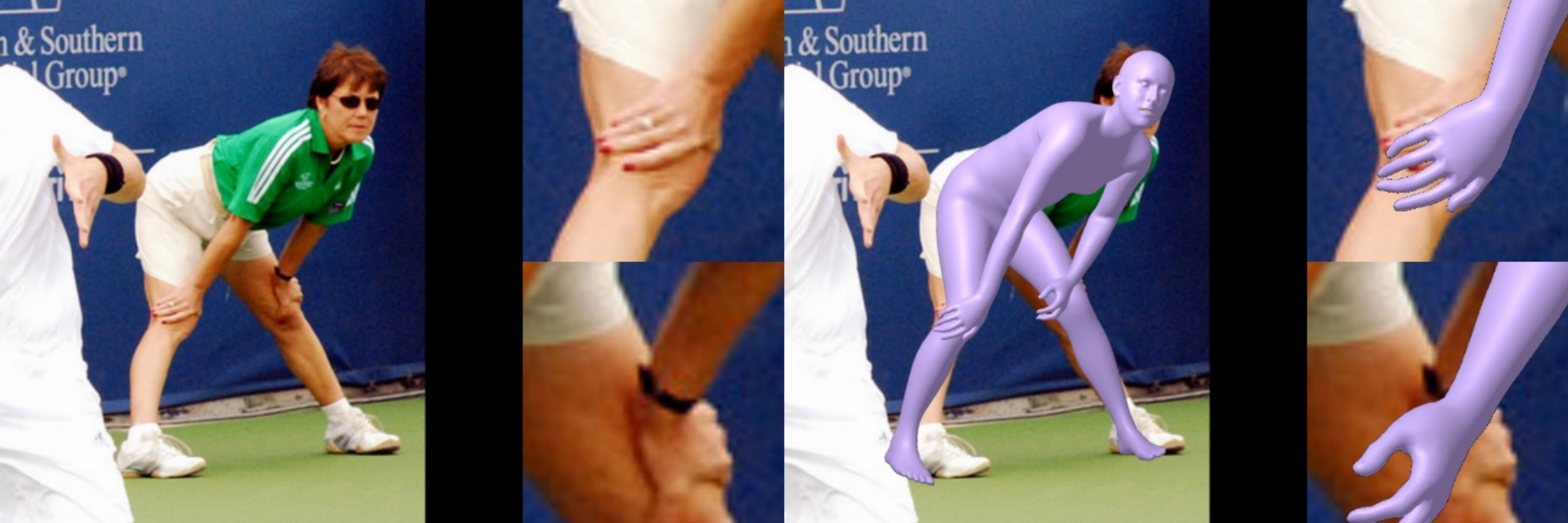}
		\caption{Challenging hand pose}
    \end{subfigure}
    \begin{subfigure}[b]{0.49\textwidth}
		\includegraphics[width=1\textwidth]{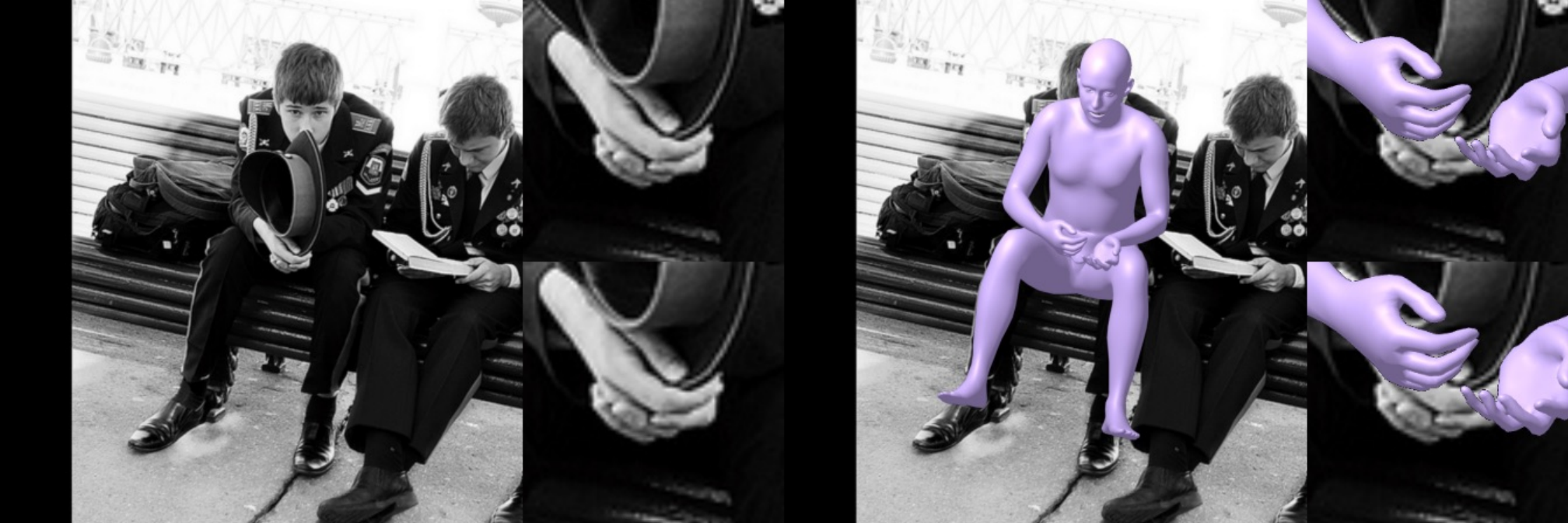}
		\caption{Interacting hands}
    \end{subfigure}
	\caption{Misaligned reconstructions of our approach. Misalignment comes from (a) inaccurate bone length (body shape), (b) inaccurate body pose, (c)(d) inaccurate hand poses under challenging hand poses, occlusions, and interactions. Samples come from the COCO validation set.
	}
	\vspace{-5mm}
	\label{fig:failed_cases}
\end{figure*}

\textbf{Failure Cases.}
Due to the rotational pose representation of the kinematic model, the full-body alignment of PyMAF-X heavily relies on the accuracy of body pose estimation. Moreover, the misalignment may also occur when the body shape is inaccurate since it affects the body bone length.
Besides, it is still challenging for PyMAF-X to handle challenging hand poses or interacting hands.
Fig.~\ref{fig:failed_cases} visualizes some erroneous results of our approach, where PyMAF-X produces misaligned results due to the issues mentioned above.

\begin{figure}[t]
	\centering
	\begin{tikzpicture}[remember picture,overlay]
	\node[font=\fontsize{8pt}{8pt}\selectfont, rotate=90] at (0,3.5) {Image};
 	\node[font=\fontsize{8pt}{8pt}\selectfont, rotate=90] at (0,1.9) {SPIN~\cite{kolotouros2019learning}};
 	\node[font=\fontsize{8pt}{8pt}\selectfont, rotate=90] at (0,0.3) {PARE~\cite{kocabas2021pare}};
	\node[font=\fontsize{8pt}{8pt}\selectfont, rotate=90] at (0,-1.4) {Baseline};
	\node[font=\fontsize{8pt}{8pt}\selectfont, rotate=90] at (0,-3.0) {PyMAF};
	\end{tikzpicture}
	\foreach \idx in {1,2,3,4,5} {
		\begin{subfigure}[h]{0.082\textwidth}
			\centering
			\foreach \sub in {0,1,2,3,4} {
    			\pgfmathsetmacro\imidx{int(\sub * 5 + \idx)}
    			\includegraphics[width=1.1\textwidth]{fig/vis/coco/coco_\imidx.pdf}
    		}
		\end{subfigure}
	}
	\vspace{-6mm}
	\caption{Qualitative comparison of reconstruction results on the COCO validation set.}
	\vspace{-2mm}
	\label{fig:smpl_cocoDemo}
\end{figure}

\begin{figure*}[t]
	\centering
	\foreach \idx in {1, 2, 3, 4, 5, 6, 7, 8, 9, 10, 11, 12, 13, 14, 15, 16, 17, 18, 19, 20, 21, 22, 23, 24, 25, 26, 27, 28, 29, 30, 31, 32, 33, 34, 35, 36} {
		\begin{subfigure}[h]{0.32\textwidth}
    		\includegraphics[width=1\textwidth]{fig/vis/smplx/demo/well-aligned_\idx.pdf}
		\end{subfigure}
	}
	\\
	\vspace{1mm}
	\caption{More examples of the full-body mesh recovery results of PyMAF-X on the COCO validation set. Best viewed zoomed-in on a color screen.}
	\vspace{-5mm}
	\label{fig:well-aligned}
\end{figure*}

